\documentclass[runningheads]{llncs}

\usepackage{eccv}

\usepackage{eccvabbrv}

\usepackage{graphicx}
\usepackage{booktabs}
\usepackage{multirow}
\usepackage[table]{xcolor} 
\usepackage[accsupp]{axessibility}  
\usepackage[T1]{fontenc}
\usepackage{lmodern}

\usepackage[pagebackref,breaklinks,colorlinks,citecolor=eccvblue]{hyperref}
\usepackage{adjustbox}
\usepackage{tcolorbox}
\usepackage{amsmath}   
\usepackage[ruled,vlined]{algorithm2e}  
\usepackage{balance}
\usepackage{orcidlink}

\makeatletter
\renewcommand*{\@fnsymbol}[1]{\ensuremath{\ifcase#1\or *\or \dagger\or \ddagger\or
    \mathsection\or \mathparagraph\or \|\or **\or \dagger\dagger
    \or \ddagger\ddagger \else\@ctrerr\fi}}
\makeatother

\begin{document}

\title{What if? Emulative Simulation with World Models for Situated Reasoning} 


\author{Ruiping Liu\inst{1}\orcidlink{0000-0001-5245-2277} \and
Yufan Chen\inst{1}\orcidlink{0009-0008-3670-4567} \and
Yuheng Zhang\inst{2}\orcidlink{0009-0007-9527-2234}\and
Junwei Zheng\inst{1,3}\orcidlink{0009-0005-4390-3044}\and
Kunyu Peng\inst{1,4}\orcidlink{0000-0002-5419-9292}\and
Chengzhi Wu\inst{1}\orcidlink{0000-0003-2186-3748}\and
Chenguang Huang\inst{5}\orcidlink{0009-0008-6463-6485}\and
Di Wen\inst{1}\orcidlink{0009-0000-1693-7912}\and
Jiaming Zhang\inst{2}\orcidlink{0000-0003-3471-328X}\and
Kailun Yang\inst{2,}\thanks{\textbf{Corresponding author:} kailun.yang@hnu.edu.cn; \textbf{First author:} ruiping.liu@kit.edu}\orcidlink{0000-0002-1090-667X}\and
Rainer Stiefelhagen\inst{1}\orcidlink{0000-0001-8046-4945}
}

\authorrunning{Ruiping Liu~\textit{et al.}}

\institute{Karlsruhe Institute of Technology \and
Hunan University \and ETH Zürich \and INSAIT, Sofia University ``St. Kliment Ohridski''\and RAI Institute}

\maketitle
\begin{center}

\includegraphics[width=\linewidth]{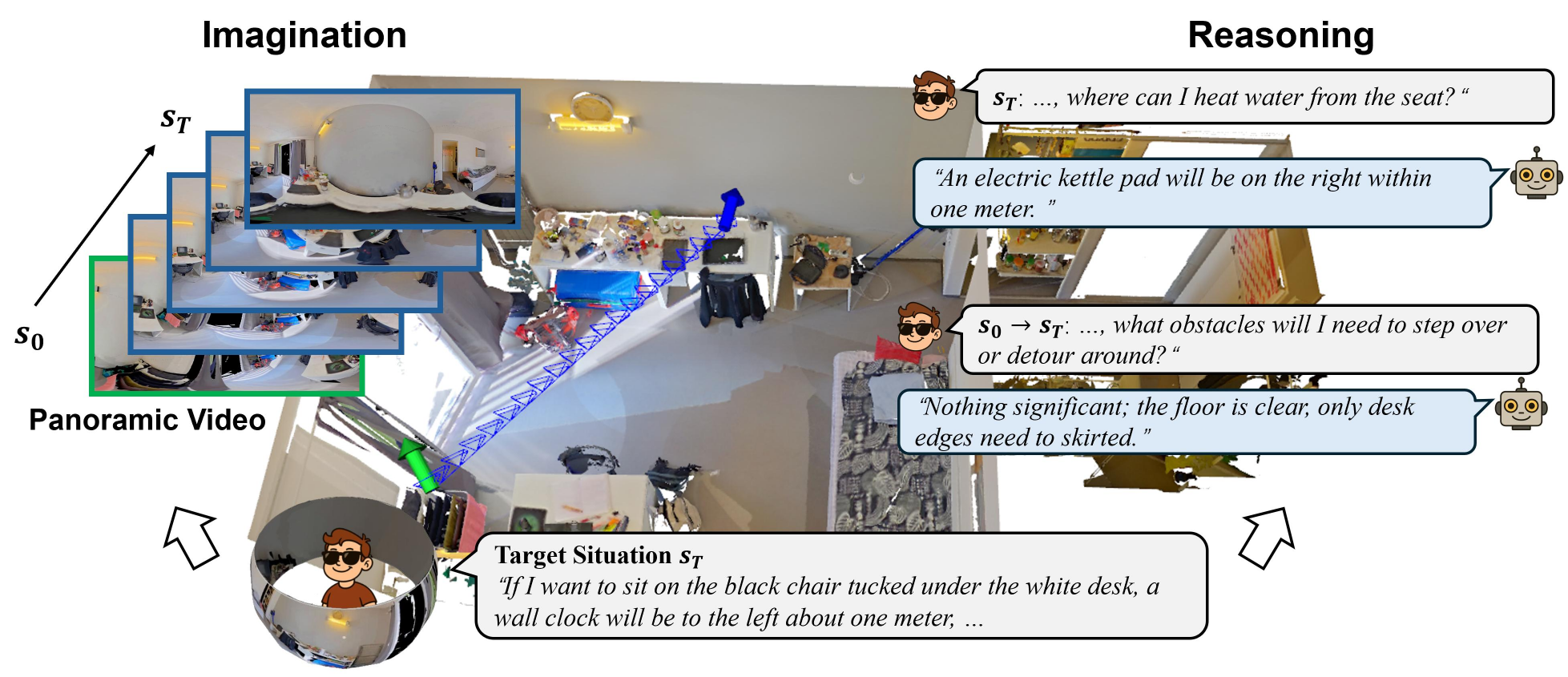}
\captionof{figure}{Emulative simulation with WanderDream. Putting oneself in the mental shoes of the agent to imagine the visual trajectory from the current perception $s_0$ toward the target situation $s_T$, and reasoning along the imagined path to answer \textit{``what-if''} questions.
Throughout the paper, \textcolor{green}{green} denotes the current state, while \textcolor{blue}{blue} represents imagination.}
\label{fig:teaser}
\end{center}
\begin{abstract}
Situated reasoning often relies on active exploration, yet in many real-world scenarios such exploration is infeasible due to physical constraints of robots or safety concerns of visually impaired users. Given only a limited observation, can an agent mentally simulate a future trajectory toward a target situation and answer spatial \textit{``what-if''} questions? We introduce WanderDream, the first large-scale dataset designed for the emulative simulation of mental exploration, enabling models to reason without active exploration. WanderDream-Gen comprises 15.8K panoramic videos across 1,088 real scenes from HM3D, ScanNet++, and real-world captures, depicting imagined trajectories from current viewpoints to target situations. WanderDream-QA contains 158K question–answer pairs, covering starting states, paths, and end states along each trajectory to comprehensively evaluate exploration-based reasoning. Extensive experiments with world models and MLLMs demonstrate (1) that mental exploration is essential for situated reasoning, (2) that world models achieve compelling performance on WanderDream-Gen, (3) that imagination substantially facilitates reasoning on WanderDream-QA, and (4) that WanderDream data exhibit remarkable transferability to real-world scenarios. The source code and all data are released at \url{https://github.com/RuipingL/WanderDream}.
\end{abstract}

\section{Introduction}
\label{sec:intro}

\begin{figure}[h!]
    \centering
    \includegraphics[width=0.6\linewidth]{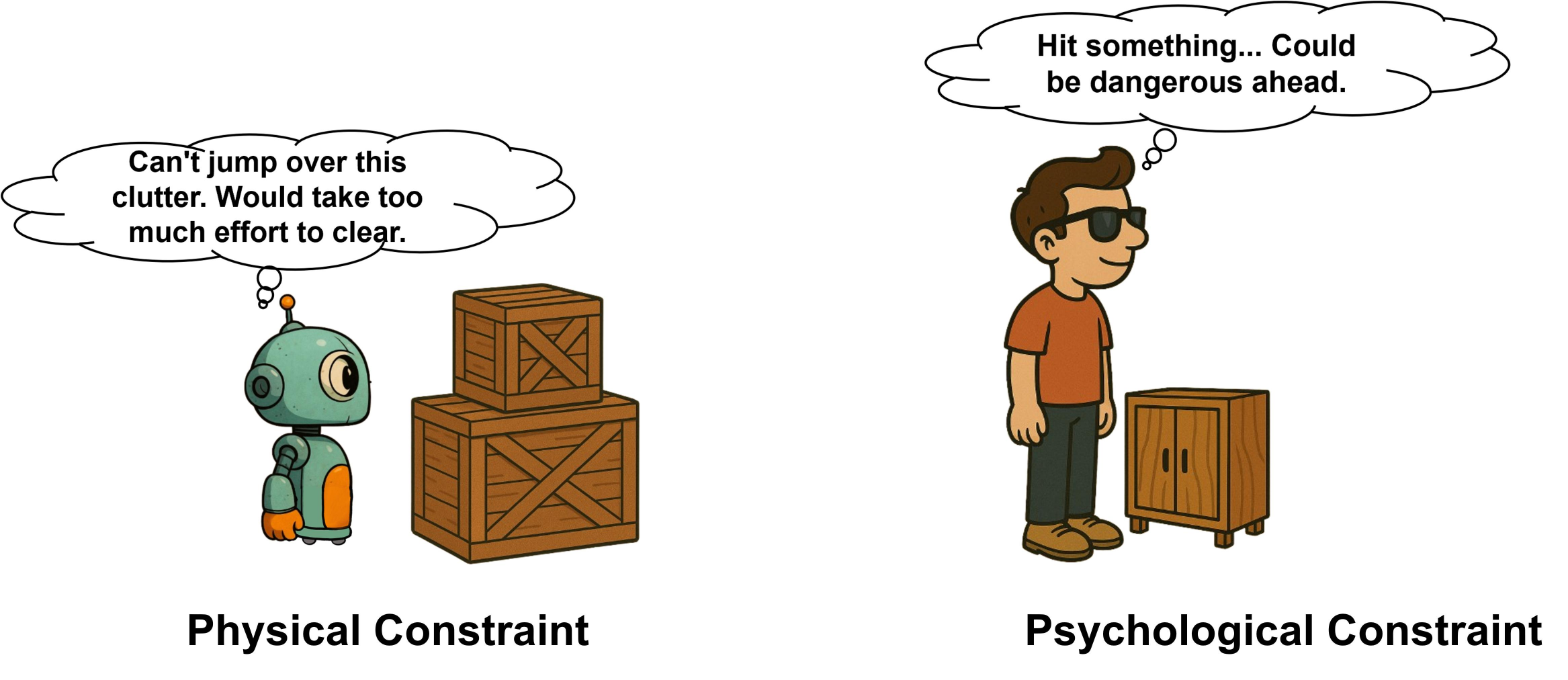}
    \caption{Constraints of active exploration: robot embodiment limits (\textit{e.g.}, inability to climb stairs) and visually impaired users’ psychological safety barriers when encountering obstacles without tactile cues.}
    \label{fig:constraints}
\end{figure}

Situated reasoning~\cite{Clancey1997SituatedCognition}, where inferences draw on the local context, the perceptual surroundings, and the grounding established through interaction~\cite{krishnaswamy-pustejovsky-2022-grounding}, is a fundamental capability of the cognitive system in both embodied agents, \textit{e.g.}, robots~\cite{embodied-reasoner, hao2024embosr, wang2025affordbot}, and body-worn assistive agents, \textit{e.g.}, wearable navigation assistants for people with visual impairments~\cite{liu2024objectfinder, liu2026not}.
However, existing situated reasoning approaches typically rely on either pre-explored scenarios~\cite{ma2022sqa3d, linghu2024multimodal_situated_3d, zhang2024spartun3d} or refinement during active exploration~\cite{yang20253d, yan2025dynamic, huang2025bye}, both of which inherently depend on physical exploration and are constrained by various real-world limitations, as illustrated in Fig.~\ref{fig:constraints}.
Due to mechanical design differences, robots are subject to distinct physical constraints. 
For example, warehouse robots can only operate within flat-grid warehouse zones, must continually adapt to new obstacles~\cite{inViaRobotics2016Overcoming}, and are unable to navigate stairs~\cite{seo2023stair} or uneven terrain~\cite{Analysis2005StairsClimbing}.
In contrast, although more flexible in their body movement, visually impaired individuals could face psychological constraints. 
They may hesitate to explore further when they feel unsafe~\cite{Mullins2019CBTBlind, Xu2025TravelAids}, \textit{e.g.}, when encountering obstacles that block their path~\cite{vanMunster2021Barriers}.
Moreover, the explore-then-understand paradigm fails in dynamic environments~\cite{liusituat3dchange}, where continuous changes demand ongoing memory updates. In such cases, imagination can bridge the gap by enabling agents to understand their potential situations based on their current egocentric view without additional physical movement. World models can serve as the imagination engine~\cite{ha2018worldmodels, lecun2022path,zhou2024robodreamer} for answering these situated \textit{``what-if''} questions.

\begin{figure}[t]
    \centering
    \includegraphics[width=0.8\linewidth]{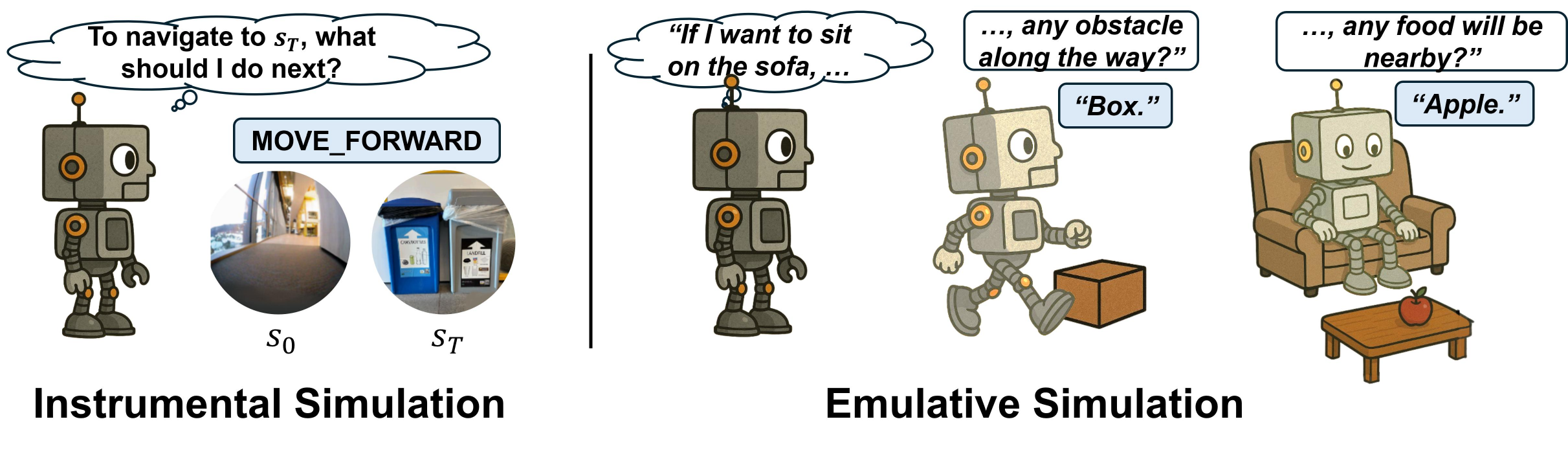}
    \caption{Two layers of mental imagination. Task-oriented instrumental simulation (left), such as Navigation World Models~\cite{bar2025navigation_world_models}, and experience-oriented emulative simulation (right), empowered by the proposed WanderDream.}
    \label{fig:simulation}
\end{figure}

Mental imagination is divided into two layers~\cite{moulton2011imagining}: instrumental simulation and emulative simulation, as illustrated in Fig.~\ref{fig:simulation}. 
Instrumental simulation is task-oriented to facilitate decision making and has been widely explored with world models, \textit{e.g.} in imagination-based navigation~\cite{koh2021pathdreamer, bar2025navigation_world_models, Nie2025WMNav, wang2023dreamwalker} and action reasoning~\cite{cen2025worldvla, zhen2024_3d_vla}. 
In contrast, \textbf{emulative simulation} is experience-oriented and serves as the core for answering \textit{``what-if''} questions by placing oneself in the mental shoes to explore the scene and reason along the path, yet it remains underexplored.
MindJourney~\cite{yang2025mindjourney} couples a world model with an MLLM for step-wise visual imagination to answer questions about view changes.
GenEx~\cite{lu2025generative} introduces a dataset of forward-moving panoramic videos to reason about the current state.
However, there is no existing dataset that provides temporally consistent video toward target situations for imaginative video generation, together with reasoning information along the path to enable emulative simulation.

\sloppy We introduce \textbf{WanderDream}, a large-scale dataset, as the first benchmark for studying emulative simulation. 
It comprises WanderDream-Gen, which provides panoramic trajectories from current viewpoints to imagined target situations, and WanderDream-QA, which contains question–answer pairs for evaluating reasoning along these imagined trajectories.
To support human–robot collaboration, world models are expected to imagine situations from both perspectives, \textit{e.g.}, when a robot aims to navigate to a chair while a human intends to sit on it.
We collect robot navigation situations from HM3D and human action situations from ScanNet++, yielding $15.8$K videos across $1{,}088$ real scenes.
For reasoning along trajectories, we design $10$ QA types covering the start state, the path, and the end state, resulting in a total of $158$K QA pairs.

\textbf{Extensive experiments} with various world models and MLLMs under different frameworks are conducted to verify that imagination is essential for situated reasoning, to evaluate the performance of world models on WanderDream-Gen, to examine how world-model-based imagination enhances reasoning on WanderDream-QA, and to assess the transferability of WanderDream data to the real world. 
The results show that although the location of the target situation can be reasoned from the current perception, imagination remains essential for situated reasoning. Moreover, world models that perform better on WanderDream-Gen tend to enable stronger reasoning on WanderDream-QA. 
Despite differences between WanderDream and real-world recordings, such as partial occlusions caused by the real agent, WanderDream exhibits strong transferability for both video generation and reasoning tasks.

\section{Related Work}
\begin{table*}[]
    \centering
    \footnotesize
        \caption{Datasets for situated reasoning. Visual quality: photographic \raisebox{-0.2em}{\includegraphics[height=0.9em]{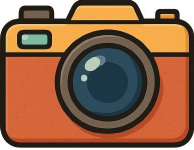}} \textit{vs.} photorealistic \raisebox{-0.2em}{\includegraphics[height=0.9em]{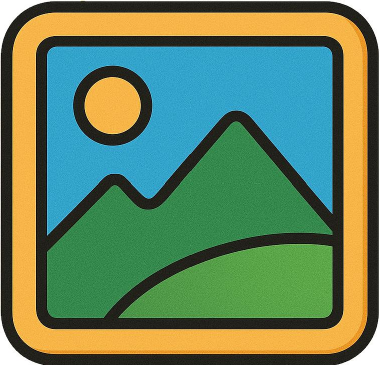}}. Task inputs and outputs include 3D point cloud \raisebox{-0.2em}{\includegraphics[height=0.9em]{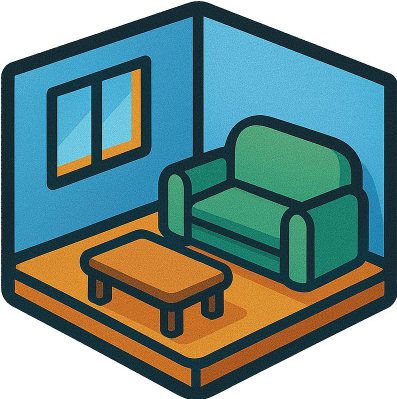}}, perspective video \raisebox{-0.2em}{\includegraphics[height=0.9em]{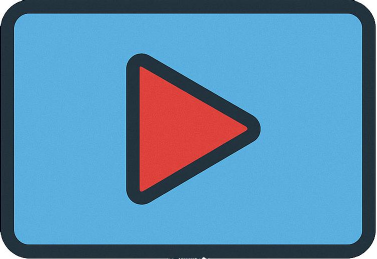}}, panoramic video \raisebox{-0.2em}{\includegraphics[height=0.9em]{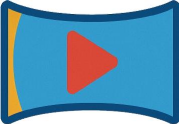}}, multi image \raisebox{-0.2em}{\includegraphics[height=0.9em]{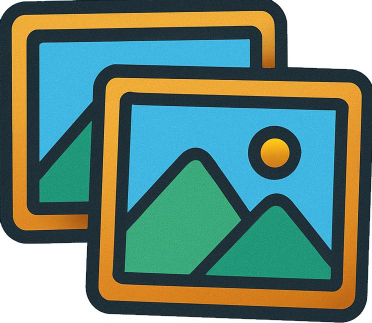}}, panoramic image \raisebox{-0.2em}{\includegraphics[height=0.9em]{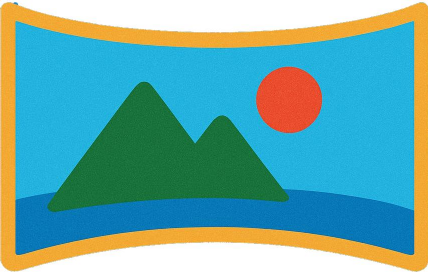}}, audio \raisebox{-0.2em}{\includegraphics[height=0.9em]{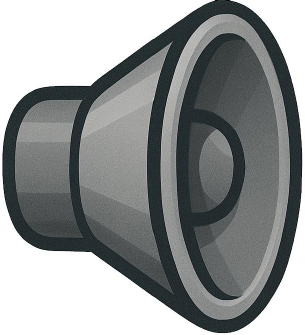}}, and text \raisebox{-0.2em}{\includegraphics[height=0.9em]{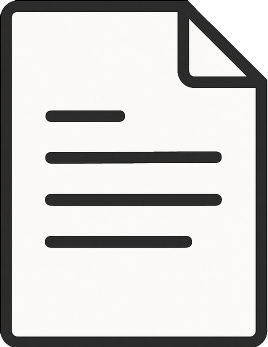}}. Visual modalities: RGB, point cloud (pcd), depth map (D), and semantic map (S).}
    \footnotesize
    \resizebox{\textwidth}{!}{
    \begin{tabular}{c c c c c c c c c c}
    \toprule
         \textbf{Dataset}& \textbf{Venue} & \textbf{Type} & \textbf{\#Scene} & \textbf{\#Video}& \textbf{\#QA Types}& \textbf{\#QA} &\textbf{Input}& \textbf{Output} & \textbf{Modalities} \\
    \midrule
        SQA3D~\cite{ma2022sqa3d}& ICLR'23& \raisebox{-0.2em}{\includegraphics[height=1em]{figures/icon_camera.pdf}}& 650&-&6&33.4K&\raisebox{-0.2em}{\includegraphics[height=1em]{figures/icon_3d.pdf}} \raisebox{-0.2em}{\includegraphics[height=1em]{figures/icon_txt.pdf}}&\raisebox{-0.2em}{\includegraphics[height=1em]{figures/icon_txt.pdf}}&RGB, pcd\\

        MSQA~\cite{linghu2024multimodal_situated_3d}& NeurIPS'24& \raisebox{-0.2em}{\includegraphics[height=1em]{figures/icon_camera.pdf}}& 1734&-&8&251K&\raisebox{-0.2em}{\includegraphics[height=1em]{figures/icon_3d.pdf}} \raisebox{-0.2em}{\includegraphics[height=1em]{figures/icon_txt.pdf}}&\raisebox{-0.2em}{\includegraphics[height=1em]{figures/icon_txt.pdf}}&RGB, pcd\\
        
        Situat3DChange~\cite{liusituat3dchange}& NeurIPS'25& \raisebox{-0.2em}{\includegraphics[height=1em]{figures/icon_camera.pdf}}& 903&-&9&121K&\raisebox{-0.2em}{\includegraphics[height=1em]{figures/icon_3d.pdf}} \raisebox{-0.2em}{\includegraphics[height=1em]{figures/icon_txt.pdf}}&\raisebox{-0.2em}{\includegraphics[height=1em]{figures/icon_txt.pdf}}&RGB, pcd\\
        
        Pano-AVOA~\cite{yun2021pano}& ICCV'21&\raisebox{-0.2em}{\includegraphics[height=1em]{figures/icon_camera.pdf}} & n/a& 5.4K& 2&51.7K&\raisebox{-0.2em}{\includegraphics[height=1em]{figures/icon_pano_video.pdf}} \raisebox{-0.2em}{\includegraphics[height=1em]{figures/icon_audio.pdf}} \raisebox{-0.2em}{\includegraphics[height=1em]{figures/icon_txt.pdf}}&\raisebox{-0.2em}{\includegraphics[height=1em]{figures/icon_txt.pdf}}&RGB\\
        
        SAT~\cite{ray2024sat}& COLM'25&\raisebox{-0.2em}{\includegraphics[height=1em]{figures/icon_camera.pdf}} \raisebox{-0.2em}{\includegraphics[height=1em]{figures/icon_photo.pdf}}&22K&-&6&175K&\raisebox{-0.2em}{\includegraphics[height=0.9em]{figures/icon_imgpair.pdf}} \raisebox{-0.2em}{\includegraphics[height=1em]{figures/icon_txt.pdf}}&\raisebox{-0.2em}{\includegraphics[height=1em]{figures/icon_txt.pdf}}&RGB \\
        
        MMSI-Bench~\cite{yang2025mmsi}&arXiv'25& \raisebox{-0.2em}{\includegraphics[height=1em]{figures/icon_camera.pdf}}&n/a&-&11&1K&\raisebox{-0.2em}{\includegraphics[height=0.9em]{figures/icon_imgpair.pdf}} \raisebox{-0.2em}{\includegraphics[height=1em]{figures/icon_txt.pdf}}&\raisebox{-0.2em}{\includegraphics[height=1em]{figures/icon_txt.pdf}}&RGB\\
        
        OpenEQA~\cite{OpenEQA2023}& CVPR'24& \raisebox{-0.2em}{\includegraphics[height=1em]{figures/icon_camera.pdf}}& 187&187&7&1.6K&\raisebox{-0.2em}{\includegraphics[height=0.9em]{figures/icon_video.pdf}} \raisebox{-0.2em}{\includegraphics[height=1em]{figures/icon_txt.pdf}}&\raisebox{-0.2em}{\includegraphics[height=1em]{figures/icon_txt.pdf}}&RGB\\
        
        VSI-Bench~\cite{yang2025thinking}& CVPR'25& \raisebox{-0.2em}{\includegraphics[height=1em]{figures/icon_camera.pdf}}& 288&288&8&5K&\raisebox{-0.2em}{\includegraphics[height=0.9em]{figures/icon_video.pdf}} \raisebox{-0.2em}{\includegraphics[height=1em]{figures/icon_txt.pdf}}&{\includegraphics[height=1em]{figures/icon_txt.pdf}}&RGB\\

        DSI-Bench~\cite{zhang2025dsibenchbenchmarkdynamicspatial}&arXiv'25&\raisebox{-0.2em}{\includegraphics[height=1em]{figures/icon_camera.pdf}}&-&1K&6&1.7K&\raisebox{-0.2em}{\includegraphics[height=0.9em]{figures/icon_video.pdf}} \raisebox{-0.2em}{\includegraphics[height=1em]{figures/icon_txt.pdf}}&\raisebox{-0.2em}{\includegraphics[height=1em]{figures/icon_txt.pdf}}&RGB\\

        VSI-590K~\cite{yang2025cambrian} & arXiv'25& \raisebox{-0.2em}{\includegraphics[height=1em]{figures/icon_camera.pdf}} \raisebox{-0.2em}{\includegraphics[height=1em]{figures/icon_photo.pdf}} & n/a&6K&8&590.7K&\raisebox{-0.2em}{\includegraphics[height=0.9em]{figures/icon_video.pdf}} / \raisebox{-0.2em}{\includegraphics[height=0.9em]{figures/icon_imgpair.pdf}} \raisebox{-0.2em}{\includegraphics[height=1em]{figures/icon_txt.pdf}}&\raisebox{-0.2em}{\includegraphics[height=1em]{figures/icon_txt.pdf}}&RGB, S\\

        \rowcolor{gray!20}
         \textbf{WanderDream (Ours)}&-&\raisebox{-0.2em}{\includegraphics[height=1em]{figures/icon_camera.pdf}} \raisebox{-0.2em}{\includegraphics[height=1em]{figures/icon_photo.pdf}} &1088&15.8K&10&158.1K&\raisebox{-0.2em}{\includegraphics[height=1em]{figures/icon_pano_img.pdf}} \raisebox{-0.2em}{\includegraphics[height=1em]{figures/icon_txt.pdf}}&
         \raisebox{-0.2em}{\includegraphics[height=1em]{figures/icon_pano_video.pdf}} 
         \raisebox{-0.2em}{\includegraphics[height=1em]{figures/icon_txt.pdf}}&RGB, D, S\\
    \bottomrule
    \end{tabular}}
    \label{tab:related_work}
\end{table*}

\noindent\textbf{Situated scene understanding.}
Prior scene understanding works focus primarily on allocentric relations~\cite{wang2025masked_point_entity, jia2024sceneverse, zhu2024scanreason, xiong2025_3ur_lmm, liu2023open}, neglecting the egocentric, situated perception of the agent. Recent efforts in situated reasoning attempt to imagine situations within pre-explored static scenarios. Datasets such as SQA3D~\cite{ma2022sqa3d}, MSQA~\cite{linghu2024multimodal_situated_3d}, Spartun3D~\cite{zhang2024spartun3d}, and SURPRISE3D~\cite{huang2025surprise3d} support egocentric reasoning in such contexts. 
Corresponding methods~\cite{man2024situational, huang2024embodied, wang2025affordbot} successfully solve these tasks, whereas HIS-GPT~\cite{zhao2025his_gpt} introduces a benchmark for reasoning about virtual scenes involving a virtual human. Other datasets, including Situat3DChange~\cite{liusituat3dchange} and SOK-Bench~\cite{wang2024sok_bench}, emphasize temporal evolution and situational change. Another line of work tackles situated reasoning through video streams acquired during active exploration~\cite{yang2025cambrian, yun2021pano, OpenEQA2023, qin2024worldsimbench, yuan2025scene_r1, zheng2025video_3d_llm, liu2026egoexomem}, either by direct perception or through representations such as code maps~\cite{yang2025thinking}, image graphs~\cite{yang20253d, hu2025changinggrounding}, online query embeddings~\cite{zhu2025move_3d}, or situational scene graphs~\cite{sugandhika2025situational_scene_graph}. 
These approaches support decision-making during exploration, \textit{e.g.}, in navigation~\cite{jin2026panonav}. 
Multi-view frameworks like AffordBot~\cite{wang2025affordbot} and Omni-View~\cite{hu2025omniview} leverage previously seen frames for gaze and affordance reasoning, whereas STAR~\cite{wu2024star} and SpatialLadder~\cite{li2025spatialladder} introduce learning curricula bridging perception and reasoning. 
ODI Bench~\cite{yang2025odi_bench} and CFPano~\cite{zhang2025omnidirectional_spatial_modeling} benchmark the understanding of panoramic imagery.
In contrast, we address situated reasoning through emulative imagination with world models, allowing agents to reason without memory from active exploration, especially in inaccessible environments. A comparison of related datasets is shown in Tab.~\ref{tab:related_work}.

\noindent\textbf{Video generation and understanding.}
Video generation models~\cite{ho2022video,blattmann2023stable} are increasingly vital for producing realistic, temporally coherent visual content in applications such as entertainment, simulation, robotics, and data augmentation. Recent advances have moved from frame-by-frame synthesis to spatio-temporally consistent generation using diffusion-based architectures~\cite{ho2022video,blattmann2023stable,xing2024dynamicrafter,wu2023tune,zhou2022magicvideo}.
Modern approaches enhance controllability~\cite{koksal2023controllable,hu2022make, wang2025spatialvid, lu2025generative}, enable multimodal conditioning~\cite{chen2025humo,li2025multimodal}, and improve real-world fidelity~\cite{hu2025simulating}, yielding high-quality, semantically aligned outputs.
Unified frameworks deal with video generation and understanding separately, and have emerged in pinhole video settings~\cite{tan2025omni,chen2024sharegpt4video, xie2025showo}, while panoramic video generation~\cite{wang2025panogen++,wang2024360dvd,wen2024panacea,liu2025dynamicscaler,xia2025panowan,gui2025image_world} is gaining traction for its immersive 360{\textdegree} views and richer scene understanding~\cite{chen2024_360x,zhang2025towards_360r1}. 
While front views are limited when facing room corners or large objects, panoramic views are widely adopted in real-world edge devices~\cite{wu2025quadreamer, wei2024onebev, song2025anomaly_detection,weiss2020navigation_agents}. 
We therefore use panoramic videos to facilitate visual imagination with a wider field of view, while also releasing single-view videos.

\noindent\textbf{World models for spatial intelligence.}
World models aim to internalize environmental dynamics, enabling agents to predict, plan, and act through imagination rather than direct perception~\cite{ding2025understanding}. 
Early work focuses on generating coherent virtual worlds~\cite{li2025worldgrow, zhou2024holodreamer, chen2025flexworld, hunyuanworld2025tencent, genie3_deepmind2025, marble_worldlabs2025} that support free user exploration.
Following the two-layer structure of imagination~\cite{moulton2011imagining}, most current world models for real-world spatial intelligence remain within instrumental simulation, which predicts successive states needed to complete a task.
Navigation World Models~\cite{bar2025navigation_world_models} and PathDreamer~\cite{koh2021pathdreamer} synthesize future views from an initial image and a sequence of given actions. Models including 3D VLA~\cite{zhen2024_3d_vla}, WorldVLA~\cite{cen2025worldvla}, OccLLaMA~\cite{wei2024occllama}, DrivingGPT~\cite{chen2025drivinggpt}, and AdaWorld~\cite{gao2025adaworld} integrate vision, language, and action. 
EgoTwin~\cite{xiu2025egotwin} predicts both egocentric views and human poses from stepwise action descriptions. Theoretical insights from Richens \textit{et al.}~\cite{richens2024robust_causal} emphasize that robust intelligence requires causal world modeling, while benchmarks such as WAGIBench~\cite{veerabadran2025benchmarking_wearable} highlight embodied goal inference in rea- world settings. In this work, we focus on the second layer, emulative simulation, which imagines the visual experience along the path to an \textit{``if''} target situation and supports answering \textit{``what-if''} questions.

\section{WanderDream}
WanderDream supports emulative simulation through two components: Wander-Dream-Gen (Sec.~\ref{subsec:wanderdream-gen}) for spatial imagination from the current state to the target situation, and WanderDream-QA (Sec.~\ref{subsec:wanderdream-qa}) for reasoning along the trajectory. Data quality is ensured through rigorous control, and dataset statistics are provided in Sec.~\ref{subsec:quality}. A small real-world test set (Sec.~\ref{subsec:sim2real}) is included to assess sim-to-real transfer and to support future evaluation.

\subsection{WanderDream-Gen}
\label{subsec:wanderdream-gen}
To build an imagination engine that understands both robotic and human situations for effective human-robot collaboration, we consider the distinct situations of each. Robots navigate toward landmarks to support further movement, whereas humans interact with nearby objects. Prior situated reasoning work~\cite{linghu2024multimodal_situated_3d, liusituat3dchange} uses an anchor object to locate the situation, and Epstein \textit{et al.}~\cite{Epstein2017CognitiveMap} report that imagined trajectories within the cognitive map follow the shortest path. Following these principles, we treat object navigation as \textit{robotic situated path imagination} on HM3D~\cite{ramakrishnan2021habitat} and spatial shortest path prediction as \textit{human situated path imagination} on ScanNet++~\cite{yeshwanth2023scannet++}.

\begin{figure*}[t]
    \centering
    \includegraphics[width=\linewidth]{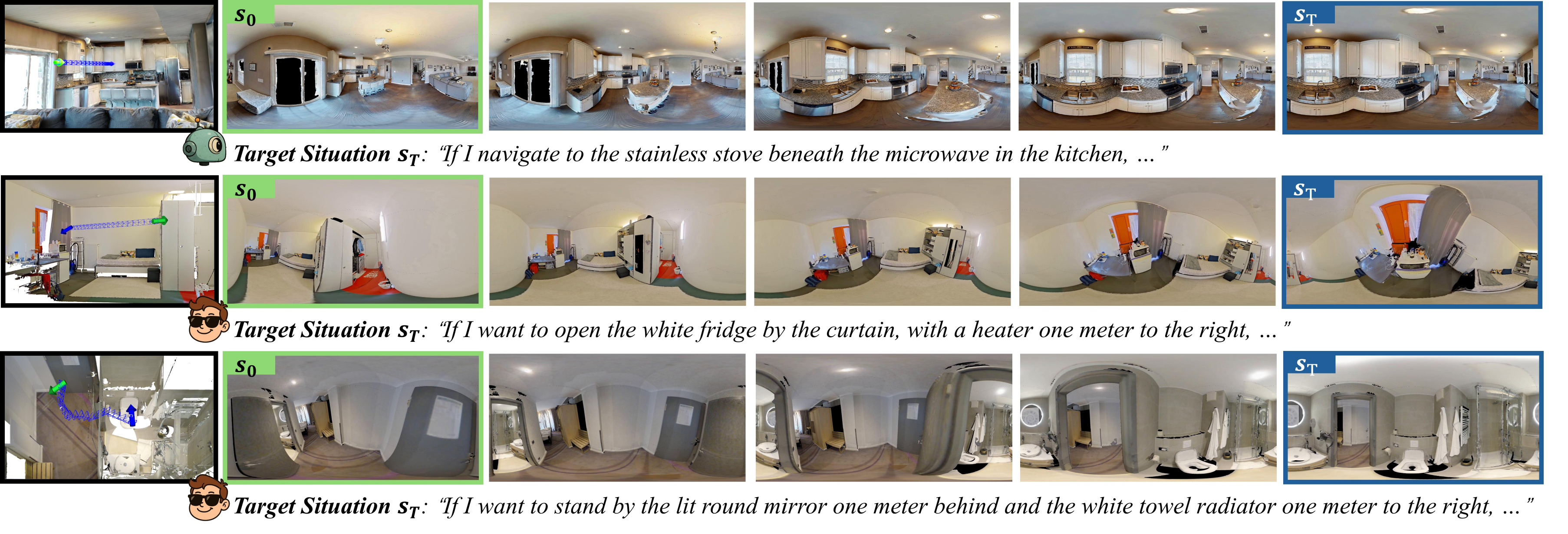}
    \caption{WanderDream-Gen. Top row: Object navigation as robotic situated path imagination in HM3D. Middle row: Human-situated perspective with direct interpolation as the shortest path when no non-traversable obstacles are present. Bottom row: Human-situated perspective with computed shortest path accounting for non-traversable obstacles (\textit{e.g.}, walls).}
    \label{fig:wanderdream_gen}
\end{figure*}

\noindent\textbf{Robotic situated path imagination.} 
For robotic agents, we select salient landmarks from 19 classes and generate ground-truth object-navigation paths using Habitat-Sim's default shortest-path planner~\cite{habitat19iccv}, discretized into actions: \texttt{MOVE\_FORWARD} ($0.25m$), \texttt{TURN\_LEFT} ($10^\circ$), and \texttt{TURN\_RIGHT} ($10^\circ$), as illustrated in the first row of Fig.~\ref{fig:wanderdream_gen}. 
The maximum path length is $5m$.

\noindent\textbf{Human situated path imagination.} 
For human situation, we adopt situation definitions from~\cite{linghu2024multimodal_situated_3d, liusituat3dchange}, including: \texttt{interacting}, \texttt{standing}, and \texttt{sitting}. 
Given human physical flexibility, such as the ability to step over obstacles like trash cans instead of navigating around them like robots, trajectory prediction becomes more complex. Due to the small room sizes in the ScanNet++ domain, a random navigable starting point is sampled at a distance of $1.5m{\sim}3m$ for each situation.
When no non-traversable obstacles exist, we employ direct path interpolation for both position and orientation as ground truth (middle row of Fig.~\ref{fig:wanderdream_gen}). 
To mimic the shortest path when non-traversable obstacles are present, we build a 3D Probabilistic Roadmap (PRM) and run Dijkstra’s algorithm on capsule-validated edges with vertical penalties. 
Invalid segments are revalidated and locally repaired using micro-PRM
(bottom row of Fig.~\ref{fig:wanderdream_gen}).
To mimic the distortion characteristics of real-world egocentric panoramas, we apply random head pitch variations within $\pm30^{\circ}$ at the start frame. Since the end state is imagined, the view is kept horizontally aligned for \texttt{standing} and \texttt{sitting} situations, while during \texttt{interacting}, the head is oriented toward the target object with pitch constrained within ${\pm}30^{\circ}$.
In addition to the RGB videos, the corresponding depth maps, semantic maps, and camera poses will also be released.

\subsection{WanderDream-QA}
\label{subsec:wanderdream-qa}

\noindent\textbf{QA categories.}
For each video, exactly ten questions are distributed along the trajectory, which is divided into three phases: the start state ($s_0$) with three questions, the path phase ($s_0{\rightarrow}s_T$) with four questions, and the end state ($s_T$) with three questions. Each question belongs to a specific reasoning category, inspired by existing situated reasoning benchmarks~\cite{lu2025generative, ray2024sat, liu2025navr1, OpenEQA2023, jia2025omnispatial}.

At the start state, we evaluate \textit{Object Awareness}, focusing on nearby objects within a local range to help localize the agent and understand its immediate surroundings. 
\textit{Navigability Reasoning} assesses whether paths are clear in the four cardinal directions, whereas \textit{Ego-Target Orientation} captures the overall spatial relationship between the agent and the target situation.

During the path phase, \textit{Landmark Sequencing} evaluates the order in which landmarks are encountered along the trajectory. 
\textit{Spatial Estimation} measures the path length to estimate the effort needed to reach the target. 
\textit{Obstacle Reasoning} identifies obstacles that need to be stepped over by humans and those block robot movement.
For robots, 
\textit{Route Planning} determines the necessary turning and forward movements. For humans, where path planning is less natural, we INSTEAD consider \textit{Relative Distance Change}, indicating whether the agent is moving closer to or farther from a specific object along the route to the target.

At the end state, \textit{Affordance} determines whether functional objects are present near the target situation. \textit{Egocentric Spatial Relationship} describes the direction and distance of specific objects relative to the agent’s final position, and \textit{Object Proximity} compares the relative distances between pairs of objects from the agent’s viewpoint.
\begin{figure}[t]
    \centering
    \includegraphics[width=0.75\linewidth]{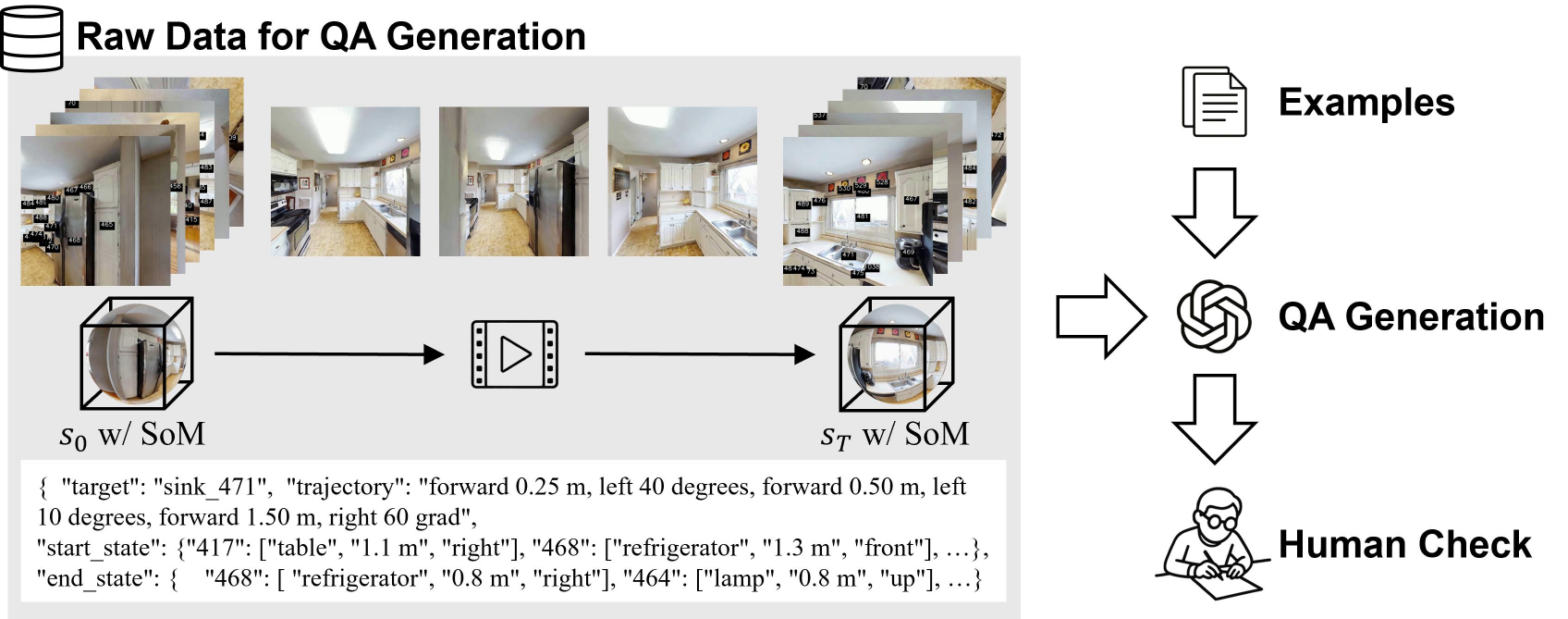}
    \caption{WanderDream-QA generation pipeline with an example in HM3D.}
    \label{fig:qa_generation}
\end{figure}

\noindent\textbf{QA generation.}
We follow established strategies~\cite{ying2025mmwalk, hong20233d, linghu2024multimodal_situated_3d} and use GPT-5~\cite{openai2025gpt5} to generate the large-scale WanderDream-QA, as shown in Fig.~\ref{fig:qa_generation}. We adopt ground-truth-annotated Set-of-Mark (SoM)~\cite{cheng2025sr3d, yang2023set} to help GPT-5 ground object instances in the image. Instance IDs are overlaid on cubemaps of the start and end states, and front-view images along the path are provided for orientation. 
In addition, a JSON file supplies trajectory metadata, with direction and distance at both the start and end states, to provide structured spatial context.

\subsection{Data Quality Control and Statistics}
\label{subsec:quality}
For WanderDream-Gen, we initially sample $40$ situations per HM3D scene and $10$ per ScanNet++ scene, followed by filtering. 
Following ReferIt3D~\cite{achlioptas2020referit_3d}, we exclude from anchor selection any category with more than six distractors of the same-category within a region in both subsets. In addition, we discard ScanNet++ scenes that are too small for imagination. 
 Each orientation of the target situation in ScanNet++ is double-checked by annotators, with reorientation applied to specific cases, such as adjusting the orientation of the situation \textit{`sitting on a piano chair'} to face the \textit{piano}. 
Situations whose anchor object is not visible in the semantic map of the final frame are automatically filtered out. 

For WanderDream-QA, human annotators first corrected erroneous answers, \textit{e.g.}, those with instance IDs or redundant phrases. 
Then, $80$ situations with $800$ QA pairs were sampled and rated for situation quality, question quality, and answer accuracy on a Likert scale ($1$ to $5$), yielding averages of $4.83{\pm}0.38$, $4.88{\pm}0.35$, and $4.73{\pm}0.47$, respectively, indicating high textual quality. 
The annotators were independent of this work and will be acknowledged.

The dataset statistics are shown in Tab.~\ref{tab:statistics}. 
We fix the video length to $21$ frames following the $4N{+}1$ convention~\cite{kong2024hunyuanvideo, wan2025wan}. 
The equirectangular panoramas have a resolution of $1024{\times}2048$.

\begin{table*}[t]
\centering
\caption{Dataset statistics of WanderDream. Trajectory length is in\textit{ meters}, and text length is in \textit{words}. Real recordings from two scenes are not included in the table.}
\resizebox{\textwidth}{!}{
\begin{tabular}{l|ccccc|ccc}
\toprule
\multirow{2}{*}{\textbf{Subset}} 
 & \multicolumn{5}{c|}{\textbf{WanderDream-Gen}} 
 & \multicolumn{3}{c}{\textbf{WanderDream-QA}} \\ 
\cmidrule(lr){2-6} \cmidrule(lr){7-9}
 & Train scenes & Train situation videos & Val. scenes & Val. situation videos & Avg. trajectory length 
 & Avg. situation length & Avg. question length & Avg. answer length \\
\midrule
ScanNet++& 856&8274&50&478& $2.19\pm0.58$&$22.29\pm2.61$&$8,81\pm1.52$&$12.19\pm3.93$\\
HM3D&144&5632&36&1404&$2.62\pm0.84$&$14.98\pm1.99$&$8.96\pm1.70$&$13.37\pm3.48$\\
\midrule
Overall&1000&13906&86&1882&$2.38\pm0.74$&$19.03\pm4.33$&$8.88\pm1.60$&$12.72\pm3.78$\\
\bottomrule
\end{tabular}}
    \label{tab:statistics}
\end{table*}
\subsection{Real-World Test Set}
\label{subsec:sim2real}
To evaluate the sim-to-real transferability of WanderDream data and to support future studies on emulative simulation, we recruited a human explorer who wore a panoramic head-mounted camera to record $26$ videos in real environments, including an office and an apartment with living room and kitchen areas. 
To focus on the role of imagination in facilitating reasoning, we generated $7$ questions per trajectory, covering the path $s_0{\rightarrow}s_T$ and end states $s_T$, resulting in $182$ QA pairs in total, which were subsequently refined by human annotators. The scale is comparable to the real-world test set in SAT~\cite{ray2024sat} ($150$ QA pairs) for a related task. Videos are also resampled to $21$ frames.

\section{Frameworks and Metrics for Emulative Simulation}
\label{sec:framework}
\begin{figure*}[t]
    \centering
    \includegraphics[width=\linewidth]{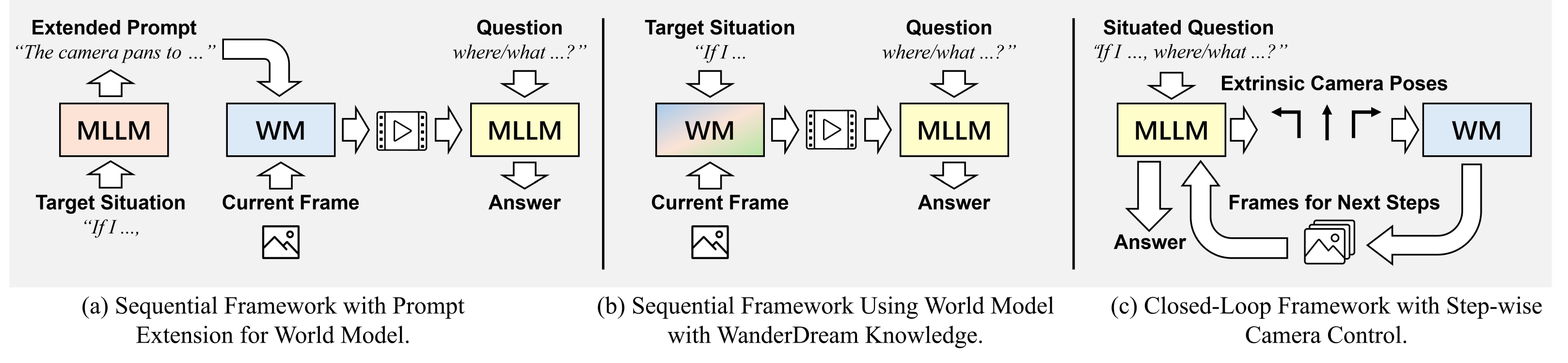}
    \caption{Frameworks for emulative simulation. (a) and (b) are the sequential frameworks we use to imagine a consistent view trajectory. (c) is the closed-loop framework of MindJourney~\cite{yang2025mindjourney}, which generates novel views step by step. \textcolor{yellow}{Yellow} MLLM modules are for reasoning, while the \textcolor{pink}{pink} one is for prompt extension. }
    \label{fig:frameworks}
    \vspace{-0.4cm}
\end{figure*}

\noindent\textbf{Sequential frameworks for trajectory simulation.} As no open-source unified model can take a question and an image and jointly output a video sequence and an answer, we design a sequential framework that combines a world model and an MLLM for imagination and reasoning. We adopt leading world models in V-Bench-2.0~\cite{zheng2025vbench}, including HunyuanVideo~\cite{kong2024hunyuanvideo}, CogVideoX~\cite{yang2024cogvideox}, and Wan~\cite{wan2025wan}, but their foreground-centered training causes zero-shot queries for target situations to produce static frames. To enable controlled camera motion, we use MLLM-driven prompt-extension scripts to decompose target trajectories into action sequences (Fig.~\ref{fig:frameworks}a), a strategy widely used in both open-source~\cite{wan2025wan, yang2024cogvideox} and commercial video generation models~\cite{openai2024sora, pika2024}. We also fine-tune models on WanderDream (Fig.~\ref{fig:frameworks}b). For reasoning, we use Qwen3-VL-32B~\cite{yang2025qwen3} and LLaVA-OneVision-1.5-8B~\cite{an2025llava}, which perform competitively on Video-MME~\cite{fu2025video}. These sequential frameworks achieve implicit camera pose control toward the target situation and answer questions along the trajectory.

\noindent\textbf{Closed-loop framework for step-wise simulation.} 
In contrast to our sequential framework that imagines the entire trajectory at once, MindJourney~\cite{yang2025mindjourney} performs emulative simulation per question through explicit step-wise camera control. 
An MLLM interprets each situated question and
proposes successive camera actions in a closed loop
(Fig.~\ref{fig:frameworks}c). Because it imagines per question via test-time scaling rather than per situation, we evaluate it only on the real-world test set. Its imagination module uses the front-view novel-view synthesis model SVC~\cite{zhou2025stable}, which fails in occluded or corner regions. To adapt it to our panoramic setting, we decompose panoramas into directional views and let the framework select the most informative view for reasoning.

\noindent\textbf{Metrics.}
We carefully select video generation metrics to evaluate imagination in WanderDream-Gen: FVD for trajectory coherence, End-FID for target-state prediction accuracy, and Spherical SSIM and LPIPS for geometric and perceptual consistency between generated videos and ground truth. Since our work focuses on imagination rather than navigation, we do not leverage traditional navigation metrics (\textit{e.g.}, path-length- or -weighted metrics). While the shortest-path assumption is suitable for cognitive map modeling, it does not necessarily reflect subjective real-world trajectories, for which no definitive ground truth exists. 

To evaluate the long-form questions on WanderDream-QA, we adopt an LLM-as-a-judge~\cite{li2025generation} evaluation to assess factual correctness. Following prior works~\cite{linghu2024multimodal_situated_3d, liusituat3dchange, ying2025mmwalk}, we use a GPT-based scoring framework 
to uniformly rate model responses:
{\setlength{\abovedisplayskip}{4pt}
 \setlength{\belowdisplayskip}{4pt}
 \begin{equation}
 C = \frac{1}{N} \sum_{i=1}^{N} \frac{s_i - 1}{4} \times 100\%,
 \end{equation}
}
where \( C \) is the overall correctness over \( N \) samples, and \( s_i \in [1, 5] \) is the GPT-assigned rating given the question, ground-truth answer, and model response.
Human evaluation over $800$ QA pairs closely aligns with GPT, with a very strong Spearman correlation of $0.9722$. Additional ablations supporting the rationale for the metric are provided in the supplementary material.

\section{Experiments}
We first address emulative simulation, where agents imagine the mental journey toward a target situation and reason about \textit{``what-if''} questions along the path. Our experiments investigate: (1) whether answering \textit{``what-if''} questions requires imagination; (2) how world models perform in imagining view trajectories toward target situations on WanderDream-Gen; (3) how imagination facilitates reasoning on WanderDream-QA; and (4) the transferability of WanderDream to real-world data.

\subsection{Implementation Details}

For fine-tuning to inject WanderDream knowledge (Fig.~\ref{fig:frameworks}b), we train videos at a resolution of \(384{\times}768\), limited by available computational resources. Some models require fixed aspect ratios, resulting in outputs smaller than \(384{\times}768\). For consistent evaluation on WanderDream-Gen, all videos are resized to \(256{\times}512\). The generated videos used for reasoning are sampled every five frames (\(s_{\Delta 5}\)) to reduce biases introduced by different video-processing pipelines in MLLMs. The world models (Sec.~\ref{sec:framework}) are fine-tuned on both subsets of WanderDream-Gen for \(8\) epochs using LoRA. For CogVideoX, LoRA is insufficient due to its image–preprocessing pipeline, so we apply supervised fine-tuning (SFT) for \(10\) epochs instead. As CogVideoX uses an \(8N{+}1\) frame scheme with \(N{=}3\), we remove four redundant frames during evaluation on WanderDream-Gen to maintain temporal alignment. All other training settings follow their original implementations and are detailed in the supplementary material. 

For all frameworks, we use Qwen3-VL as the reasoning module unless otherwise stated. MLLMs are used in a few-shot setting to answer situated questions.

\begin{figure*}[t]
    \centering
    \includegraphics[width=
    \linewidth]{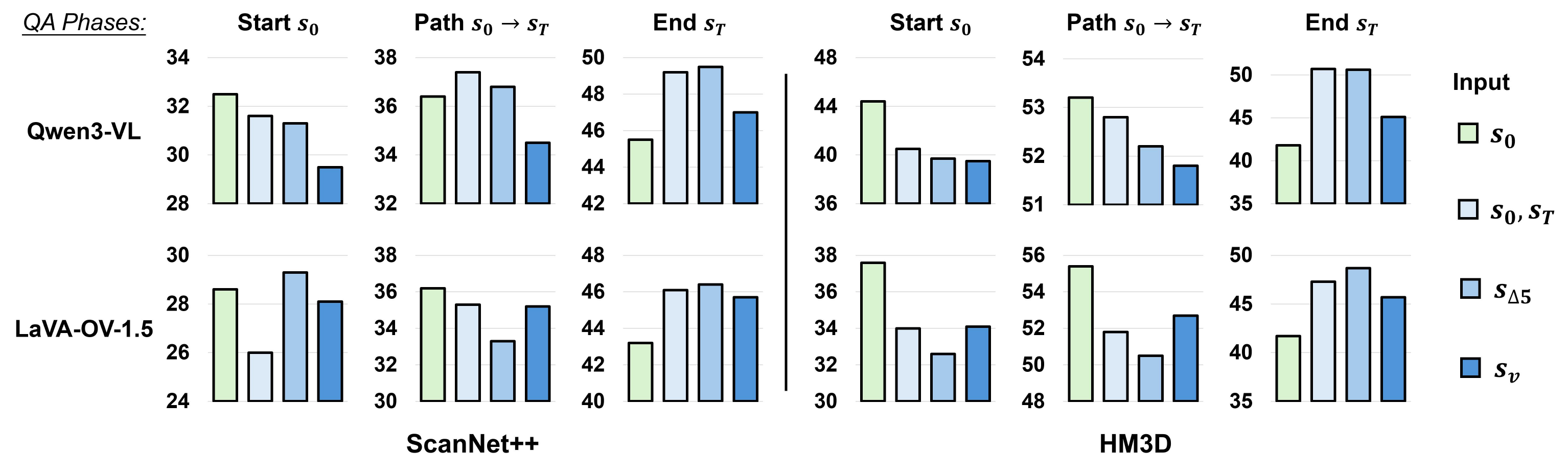}
    \caption{Results of MLLMs under different video input settings for answering questions across the phases of WanderDream-QA, used to verify the necessity of imagination for reasoning along the trajectory toward the target situation.}
    \label{fig:rq1}
\end{figure*}
\begin{table*}[t]
    \centering
    \footnotesize
    \caption{Results on WanderDream-Gen. * indicates results may be affected by the removal of redundant frames.}
    \resizebox{\textwidth}{!}{
    \begin{tabular}{c|c|cccc|cccc}
    \toprule
        \multirow{2}{*}{Model} & \multirow{2}{*}{Setting} & 
        \multicolumn{4}{c|}{ScanNet++} & 
        \multicolumn{4}{c}{HM3D} \\
        \cmidrule(lr){3-6} \cmidrule(lr){7-10}
        &&FVD $\downarrow$&End-FID $\downarrow$&S-SSIM $\uparrow$&LPIPS $\downarrow$&FVD $\downarrow$&End-FID $\downarrow$&S-SSIM $\uparrow$&LPIPS $\downarrow$\\
    \midrule
        Wan2.2 & + PE &18.73&87.02&$0.41\pm0.10$&$0.57\pm0.06$&17.62&46.86&$0.34\pm0.09$&$0.56\pm0.06$\\
        CogVideoX1.5& +PE&34.48*&85.57&$0.43\pm0.09$&$0.55\pm0.04$&38.41*&65.24&$0.33\pm0.08$&$0.56\pm0.05$\\
    \midrule
    HunyuanVideo& +LoRA &13.42&82.82&$\textbf{0.47}\pm0.10$&$0.52\pm0.06$&14.50&62.01&$\textbf{0.43}\pm0.09$&$0.49\pm0.06$\\
        Wan2.1 & +LoRA &\textbf{7.90}&\underline{70.56}&$\textbf{0.47}\pm0.10$&$\textbf{0.50}\pm0.06$&\textbf{5.96}&\underline{40.20}&$\underline{0.39}\pm0.09$&$\textbf{0.47}\pm0.05$\\
        Wan2.2 & +LoRA &\underline{9.67}&77.50&$0.45\pm0.09$&$\underline{0.51}\pm0.07$&\underline{6.19}&\textbf{40.13}&$0.38\pm0.08$&$\textbf{0.47}\pm0.06$\\
        CogVideoX1.5& +SFT &16.96*&\textbf{61.49}&$0.43\pm0.09$&$0.52\pm0.05$&22.16*&42.74&$0.34\pm0.08$&$0.53\pm0.05$\\
    \bottomrule
    \end{tabular}
    }
    \vspace{-0.4cm}
    \label{tab:gen}
\end{table*}
 
\begin{table*}[t]
    \centering
    \footnotesize
    \caption{Results on WanderDream-QA using ScanNet++ captured videos and human-perspective QA types. The first row shows the input with only the start-state frame $s_0$, without imagination.}
    \resizebox{\textwidth}{!}{
    \begin{tabular}{c|c|c|c|c|c|c|c|c|c|c|c|c|c|c}
    \toprule
        \multirow{2}{*}{Model}& \multirow{2}{*}{Setting}& \multicolumn{4}{c|}{Start $s_o$}& \multicolumn{5}{c|}{Path $s_0 \rightarrow s_T$} & \multicolumn{4}{c}{End $s_T$}\\
         && OA&NR&EO&Avg.& LS& SE&OR&DC&Avg.& OP& Aff.& ESR& Avg.\\
    \midrule
    -&-&\textbf{20.4}&\textbf{45.1}&\textbf{32.0}&\textbf{32.5}&\textbf{23.7}&53.0&\textbf{30.1}&38.7&36.4&60.1&39.1&37.4&45.5\\
    Wan2.2&+PE&20.0&\underline{42.8}&30.2&31.0&22.8&53.8&29.2&39.2&36.2&60.0&39.9&36.8&45.6\\

    CogVideoX1.5&+PE&\textbf{20.4}&42.1&\underline{31.5}&\underline{31.7}&23.2&\textbf{57.0}&29.3&38.5&\underline{37.0}&60.1&\underline{41.0}&36.6&45.9\\
    Wan2.2&+LoRA&19.4&40.6&30.7&30.2&23.0&54.0&28.7&\underline{41.0}&36.7&60.2&40.3&37.3&45.9\\
    Wan2.1&+LoRA&18.1&42.0&29.1&29.7&23.1&\underline{56.0}&28.3&39.3&36.7&\textbf{60.8}&38.0&\textbf{39.7}&\underline{46.2}\\
    HuyuanVideo&+LoRA&19.2&41.8&29.8&30.3&22.3&54.6&\underline{29.7}&40.4&36.8&59.4&39.4&39.2&46.0\\
    CogVideoX1.5&+SFT&19.5&42.5&29.9&30.6&\underline{23.5}&\underline{56.0}&29.6&\textbf{41.4}&\textbf{37.6}&\underline{60.5}&\textbf{42.6}&\underline{39.5}&\textbf{47.5}\\
    \bottomrule
    \end{tabular}}
    \label{tab:scannetpp_qa}
\end{table*}

\begin{table*}[t]
\footnotesize
    \centering
    \caption{Results on WanderDream-QA using HM3D-captured videos and robot-perspective QA types. The first row shows the input with only the start-state frame $s_0$, without imagination.}
    \resizebox{\textwidth}{!}{
    \begin{tabular}{c|c|c|c|c|c|c|c|c|c|c|c|c|c|c}
    \toprule
        \multirow{2}{*}{Model}& \multirow{2}{*}{Setting}& \multicolumn{4}{c|}{Start $s_o$}& \multicolumn{5}{c|}{Path $s_o \rightarrow s_T$} & \multicolumn{4}{c}{End $s_T$}\\
         && OA&NR&ED&Avg.& LS& SE&OR&RP&Avg.& OP& Aff.& ESR& Avg.\\
    \midrule
    -&-&\textbf{27.4}&\textbf{63.2}&42.5&\textbf{44.4}&76.0&53.5&\textbf{50.8}&32.5&53.2&54.8&44.5&26.0&41.8\\
    Wan2.2&+PE&24.6&60.8&\textbf{44.5}&43.3&76.6&56.1&\underline{44.7}&34.9&53.1&54.0&43.9&26.9&41.6\\

    CogVideoX1.5&+PE&\underline{26.7}&\underline{60.9}&\underline{44.4}&\underline{44.0}&77.6&57.2&45.1&33.5&\textbf{53.4}&55.1&44.4&25.5&41.7\\
    Wan2.2&+LoRA&20.3&55.8&42.1&39.4&\underline{77.8}&\underline{57.9}&41.3&\underline{35.3}&53.1&56.5&\underline{45.2}&\textbf{29.1}&\underline{43.6}\\

    Wan2.1&+LoRA&20.8&57.7&43.3&40.6&76.8&\textbf{58.2}&42.6&\textbf{35.9}&\textbf{53.4}&\underline{56.7}&45.0&28.2&43.3\\
    HunyuanVideo&+LoRA&22.3&59.3&44.4&42.0&\textbf{79.0}&55.4&44.1&33.6&53.0&56.6&43.0&28.0&42.5\\
    CogVideoX1.5&+SFT&20.8&57.7&41.9&40.1&76.4&57.8&41.6&34.1&52.5&\textbf{57.6}&\textbf{45.4}&\textbf{29.1}&\textbf{44.0}\\
    \bottomrule
    \end{tabular}}
    \vspace{-0.4cm}
    \label{tab:hm3d_qa}
\end{table*}




\subsection{Results}
\noindent\textbf{Necessity of imagination for \textit{``what-if''} reasoning. }
To investigate this, we input different sets of frames from the ground-truth videos to MLLMs:
(1) only the current frame \textcolor{green}{$s_0$};
(2) two frames representing the current and target states, \textcolor{blue}{$s_0, s_T$};
(3) frames sampled at an interval of $5$, resulting in five frames in total (\textcolor{blue}{$s_{\Delta5}$}); 
and
(4) the full video file \textcolor{blue}{$s_v$}, which may be influenced by varying frame-sampling strategies across different MLLMs.

The average WanderDream-QA scores across three trajectory phases on both subsets are shown in Fig.~\ref{fig:rq1}.
MLLMs often suffer from contextual interference~\cite{liang2025explaining} when multiple frames add irrelevant or redundant visual information that distracts them from key cues. 
Consequently, for start-state questions, using only the current frame \textcolor{green}{$s_0$} should give the highest accuracy, whereas for end-state questions, the combination \textcolor{blue}{$s_0$, $s_T$} is expected to perform best.

The first assumption holds. With our short trajectories, \textit{\textcolor{green}{$s_0$} allows the MLLMs understand the surroundings and plan toward the target, while it cannot interpret the end state.}
Notably, the second assumption does not hold. The model with \textcolor{blue}{$s_{\Delta5}$} performs on-par with or even better than the \textcolor{blue}{$s_0, s_T$} input when reasoning about the end state, which suggests that \textit{intermediate imagined frames also strengthen the understanding of the final situation.}
Overall, the importance of imagination increases along the trajectory.
         

\begin{table}[t]
    \centering
    \caption{Sim-to-real results on real-world test set. The panorama is decomposed into directional views to adapt to MindJourney.}
    \scriptsize
    \begin{tabular}{c|c|c|c|c|c}
         \toprule
         \multirow{2}{*}{Framework (World Model)}&\multicolumn{4}{c|}{Video Generation}&\multirow{2}{*}{QA}\\
         &FVD $\downarrow$&End-FID $\downarrow$&S-SSIM $\uparrow$&LPIPS $\downarrow$&\\
         \midrule
         -&-&-&-&-&38.5\\
         MindJourney~\cite{yang2025mindjourney} (SVC)&-&-&-&-&31.9\\
         Sequential (Wan2.2 +PE)&41.65&178.61&$0.36\pm0.06$&$0.57\pm0.08$&38.8\\
         Sequential (Wan2.1 +LoRA) &\textbf{27.49}&\textbf{175.98}& $\textbf{0.38}\pm 0.05$&$\textbf{0.54}\pm0.04$&\textbf{43.0}\\
         \bottomrule
         
    \end{tabular}
    \label{tab:sim2real}
\end{table}
\noindent\textbf{Performance of world models on WanderDream-Gen.} 
Tab.~\ref{tab:gen} summarizes the performance of world models that incorporate target situations either through prompt extension or through fine-tuning (LoRA or SFT) on the WanderDream-Gen validation set. Wan2.1 achieves the best overall performance, with strong end-state estimation (End-FID).
CogVideoX1.5 and Wan2.2, after fine-tuning on WanderDream, provide the leading end-state estimation on HM3D and ScanNet++, respectively.
Wan2.2 with camera control via prompt extension performs particularly well on HM3D, delivering superior video quality and end-state prediction without prior knowledge, outperforming fine-tuned CogVideoX1.5 in FVD ($17.62$ compared with $22.16$) and HunyuanVideo in End-FID ($46.86$ compared with $62.01$).

\noindent\textbf{Impact of imagination on WanderDream-QA reasoning.}
Tab.~\ref{tab:scannetpp_qa} and Tab.~\ref{tab:hm3d_qa} show the reasoning results on WanderDream-QA. 
The videos generated by the world models are uniformly sampled to construct \textcolor{blue}{$s_{\Delta5}$}, while retaining the start and end frames.
Although the sole start frame input \textcolor{green}{$s_0$} remains dominant for start-state reasoning due to contextual interference, \textit{we observe that models with higher video generation quality also provide stronger support for reasoning along the trajectory.}
CogVideoX1.5 trained with SFT, which produces accurate end states on the ScanNet++ subset ($61.49$ End-FID) and competitive results on HM3D ($42.74$ End-FID), achieves the highest path reasoning score ($37.6$) and end-state reasoning score ($47.5$) on ScanNet++, as well as strong end-state reasoning ($44.0$) on HM3D.
Wan2.1 trained with LoRA, which maintains consistently high video quality across all metrics, obtains the highest path reasoning score ($53.4$) on HM3D and the second-highest end-state reasoning score ($46.2$) on ScanNet++.
Wan2.2 also demonstrates strong generation and reasoning performance at the end state on HM3D.

\begin{figure}[t]
    \centering
    \includegraphics[width=\linewidth]{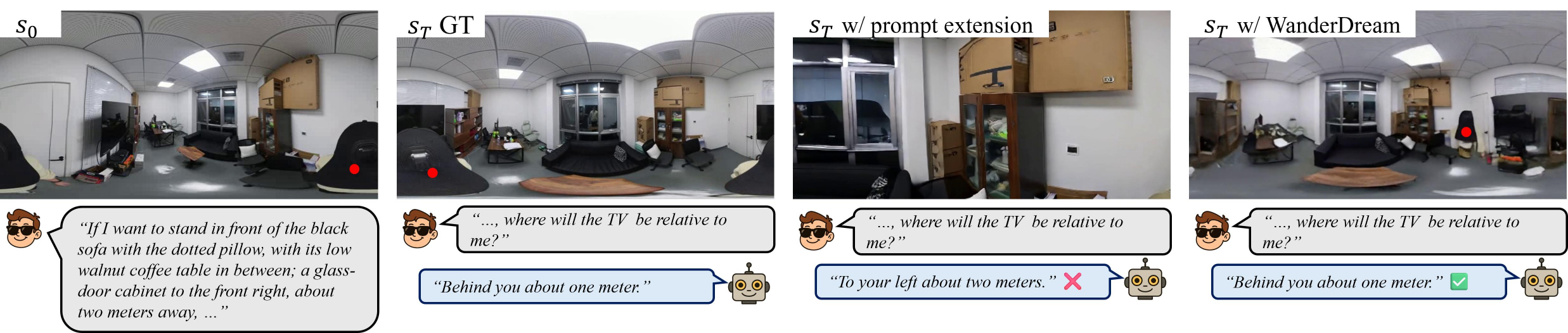}
    \caption{Sim-to-real qualitative results. Red dots ({\textcolor{red}{\textbullet}}) indicate the position of the explorer and the generated artifacts caused by it.}
    \label{fig:sim2real}
    \vspace{-0.5cm}
\end{figure}
\noindent\textbf{Transferability of WanderDream to real-world data.}
We use Wan2.1 fine-tuned with LoRA and Wan2.2 with prompt-extended camera control for the sim-to-real video generation and QA. MindJourney, a closed-loop framework that imagines and reasons step by step without consistent video generation, is evaluated only for QA.

The results in Tab.~\ref{tab:sim2real} show that Wan2.1, fine-tuned on WanderDream, surpasses Wan2.2 with prompt extension in both video generation and QA. Although the gains in End-FID, S-SSIM, and LPIPS are marginal, fine-tuning brings a clear improvement in overall video quality (FVD) together with a ${+}4.2\%$ increase in QA accuracy. Surprisingly, even though real human motion in captured videos does not follow the shortest path, training on imagined shortest-path trajectories in WanderDream still produces a notable FVD improvement that better mimics video dynamics. The real-world FVD is worse than on the WanderDream validation set. We attribute this to suboptimal real-world trajectories and varying velocity along the trajectory. MindJourney performs even worse than the MLLM that receives only the start panorama.

Fig.~\ref{fig:sim2real} shows a qualitative example. 
The fine-tuned model sometimes misinterprets agents, such as \textit{generating the explorer’s hat as an object on the door}, but still produces plausible layouts and correct answers. 
In contrast, prompt extension helps to plan roughly correct directions, \textit{e.g.}, \textit{toward the window with the cabinet on the right}, but generates front-view images instead of panoramas and places the camera in the wrong position, which leads to incorrect spatial understanding.
Despite agent occlusions and discrepancies between imagined trajectories and real exploration, WanderDream shows promising transferability for imagination and reasoning on our real-world test set, while larger-scale validation across more diverse environments is left for future work.
\section{Limitations and Future Work}
\noindent\textbf{Failure Analysis: Video Generation for Reasoning.} While more qualitative results are provided in the supplementary, Fig.~\ref{fig:failure} illustrates representative failure cases. Incorrect anchor localization leads to erroneous spatial relations. 
Under severe occlusion, models may still infer global layouts, but fine-grained details are lost, degrading performance on Affordance and Egocentric Spatial Relationship questions, which require awareness of specific objects.

\noindent\textbf{Latency in Foundation Models.} WanderDream imagines trajectories consistently, rather than answering per-question prompts~\cite{yang2025mindjourney}, so it could reduce latency when many queries are associated with a single trajectory. However, video-generation foundation models still incur substantial inference latency: we report per-trajectory runtimes of Wan 1.3B (35s), Wan 5B (53s), HunyuanVideo (211s), and CogVideo (283s). While efficiency is not the claim of this work, more efficient world models could further strengthen the emulative simulation setting.

\noindent\textbf{Reliability of Real-World Evaluation and Shortest-Path Supervision.} Our real-world set contains 182 QA pairs, exceeding the 150 in SAT~\cite{ray2024sat}, a widely used view-comparison benchmark. We deliberately prioritize quality over scale: from a large pool of collected videos, we apply strict filtering (\textit{e.g.}, stable lighting, clear anchor objects) to build a reliable sim-to-real testbed, while the large-scale validation set serves as our primary benchmark. For training, since ground truth for imagined trajectories is inherently unavailable, spatial shortest paths offer a controlled and reproducible supervision signal. This choice is also supported by classic mental-imagery findings: the mental scanning experiments of Kosslyn~\textit{et al.}~\cite{kosslyn1978visual} show that imagined trajectories preserve metric spatial structure such as straight-line distances, consistent with treating shortest paths as a proxy for imagined exploration. It is further justified empirically: as shown in Tab.~\ref{tab:sim2real}, training on WanderDream shortest-path data aligns substantially better with real-world paths than prompt extension (FVD 27.49 \textit{vs.} 41.65), demonstrating effective sim-to-real transfer.

\noindent\textbf{Toward Unified Video-and-Text Modeling.} WanderDream's emulative simulation takes the current egocentric view, a target-situation description, and a question as input, and outputs an imagined video and a corresponding answer. Existing end-to-end models~\cite{xie2025showo, xie2025showo2, wang2024emu3} typically use a unified decoder to generate either video or text, but not both. We therefore adopt a GPT-style tool-calling pipeline~\cite{openai_function_calling_guide} and provide all pipeline and evaluation prompts in the supplementary material. We encourage future work toward unified end-to-end solutions.

\begin{figure}[!t]
\footnotesize
\scriptsize
\setlength\tabcolsep{1pt}
\centering
\begin{tabular}{c c c c c c}
\raisebox{-0.5\height}{\rotatebox{90}{ScanNet++}}&
\raisebox{-0.5\height}{\includegraphics[width=0.185\textwidth]{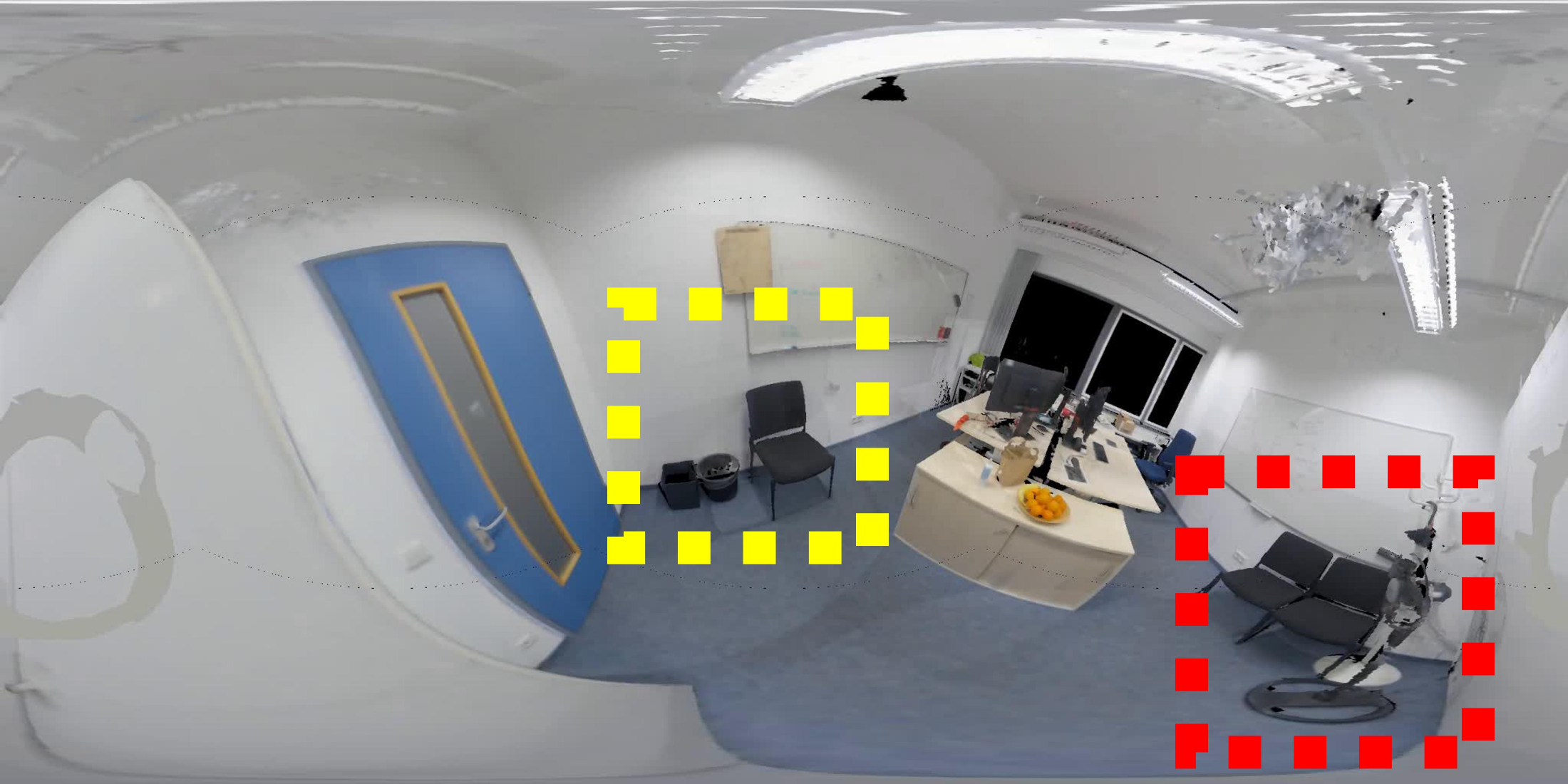}}&
\raisebox{-0.5\height}{\includegraphics[width=0.185\textwidth]{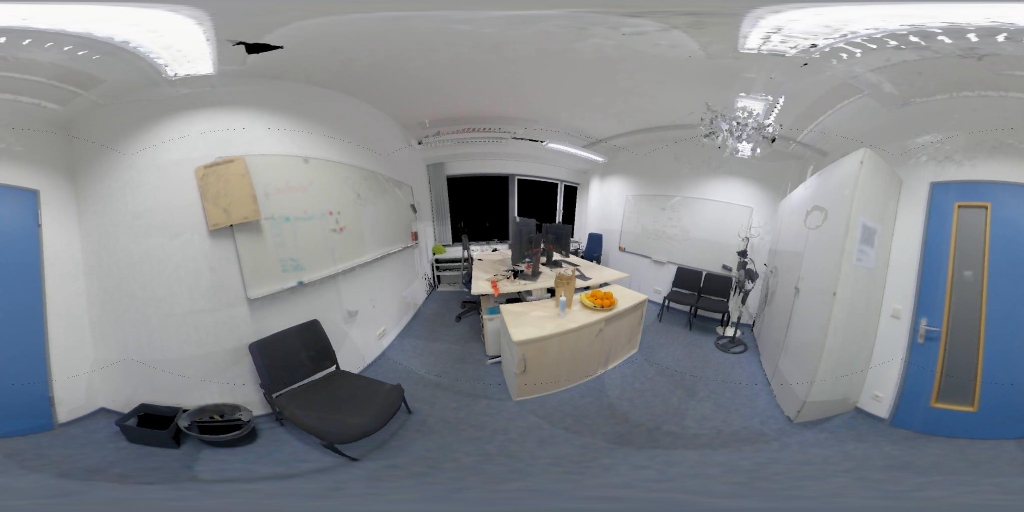}} &
\raisebox{-0.5\height}{\includegraphics[width=0.185\textwidth]{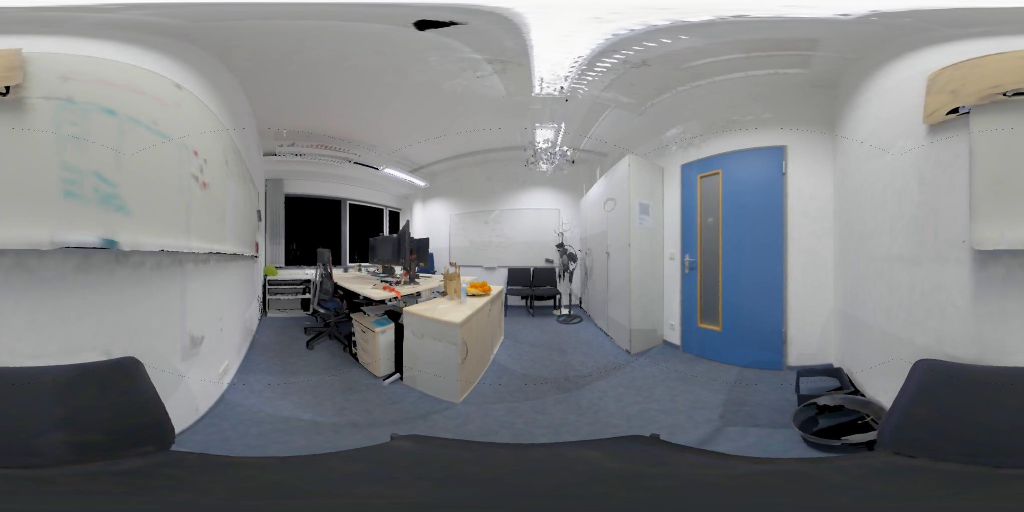}} &
\raisebox{-0.5\height}{\includegraphics[width=0.185\textwidth]{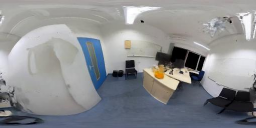}}&
\raisebox{-0.5\height}{\includegraphics[width=0.185\textwidth]{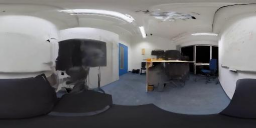}}\\
\raisebox{-0.5\height}{\rotatebox{90}{HM3D}}&
\raisebox{-0.5\height}{\includegraphics[width=0.185\textwidth]{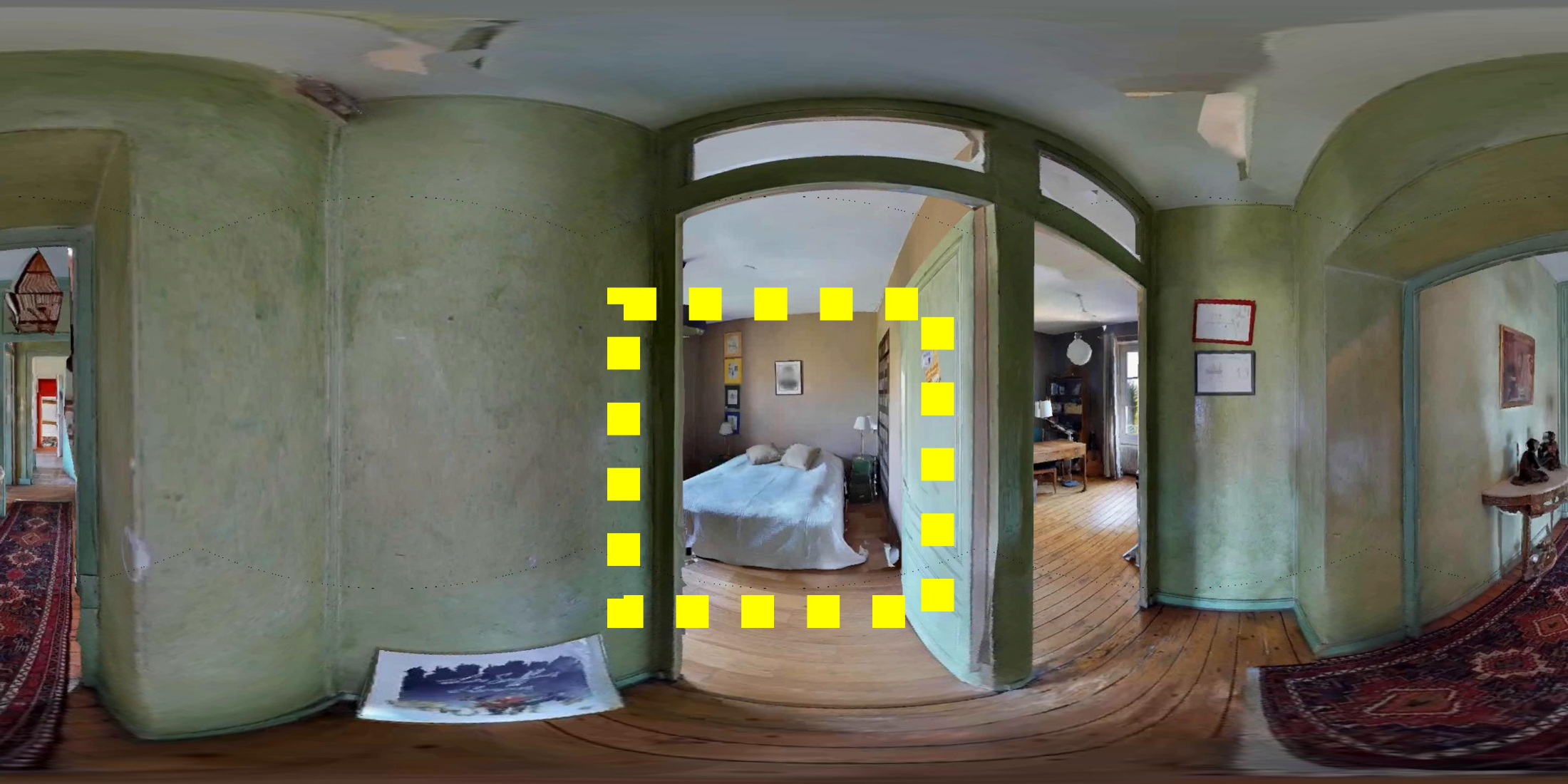}}&
\raisebox{-0.5\height}{\includegraphics[width=0.185\textwidth]{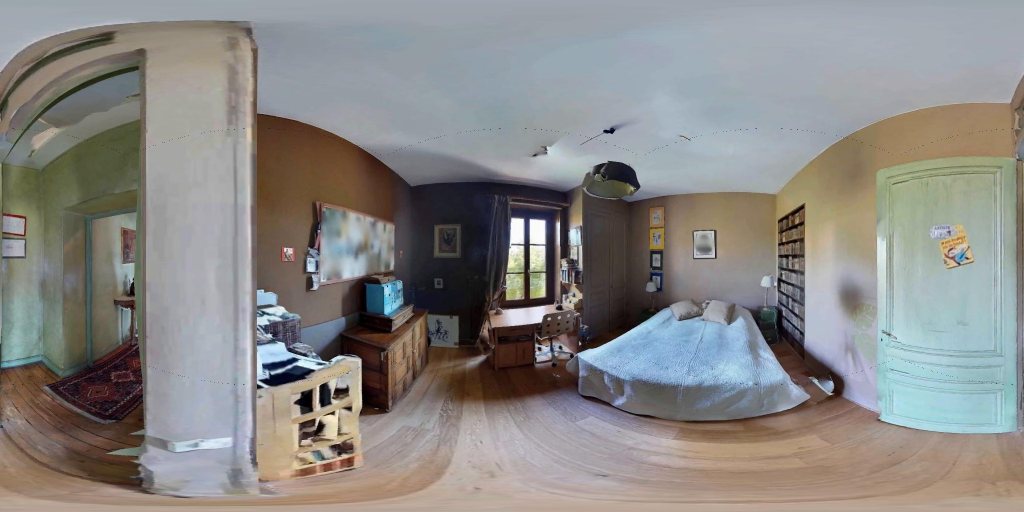}} &
\raisebox{-0.5\height}{\includegraphics[width=0.185\textwidth]{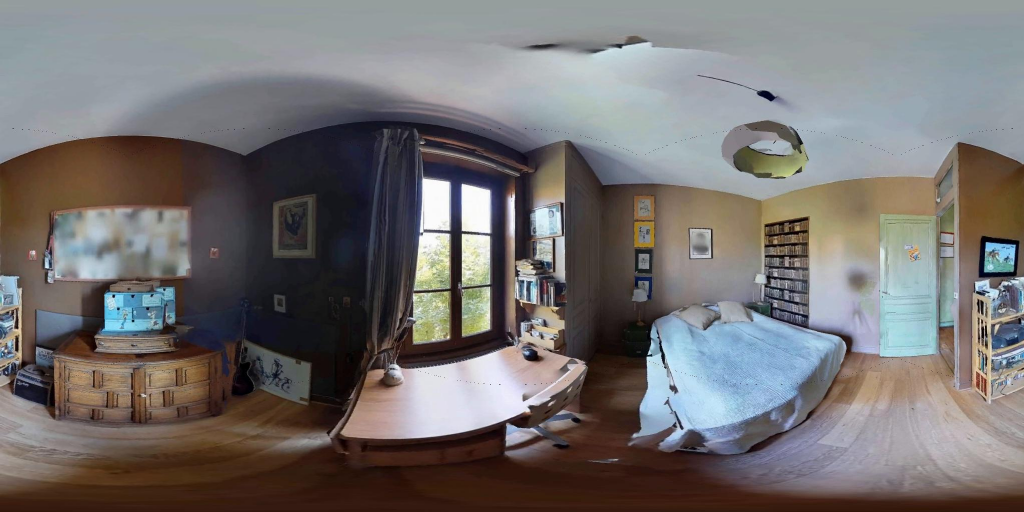}} &
\raisebox{-0.5\height}{\includegraphics[width=0.185\textwidth]{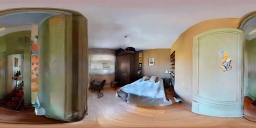}}&
\raisebox{-0.5\height}{\includegraphics[width=0.185\textwidth]{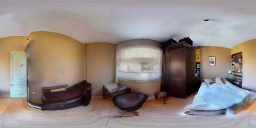}}\\
\raisebox{-0.5\height}{\rotatebox{90}{Sim2Real}}&
\raisebox{-0.5\height}{\includegraphics[width=0.185\textwidth]{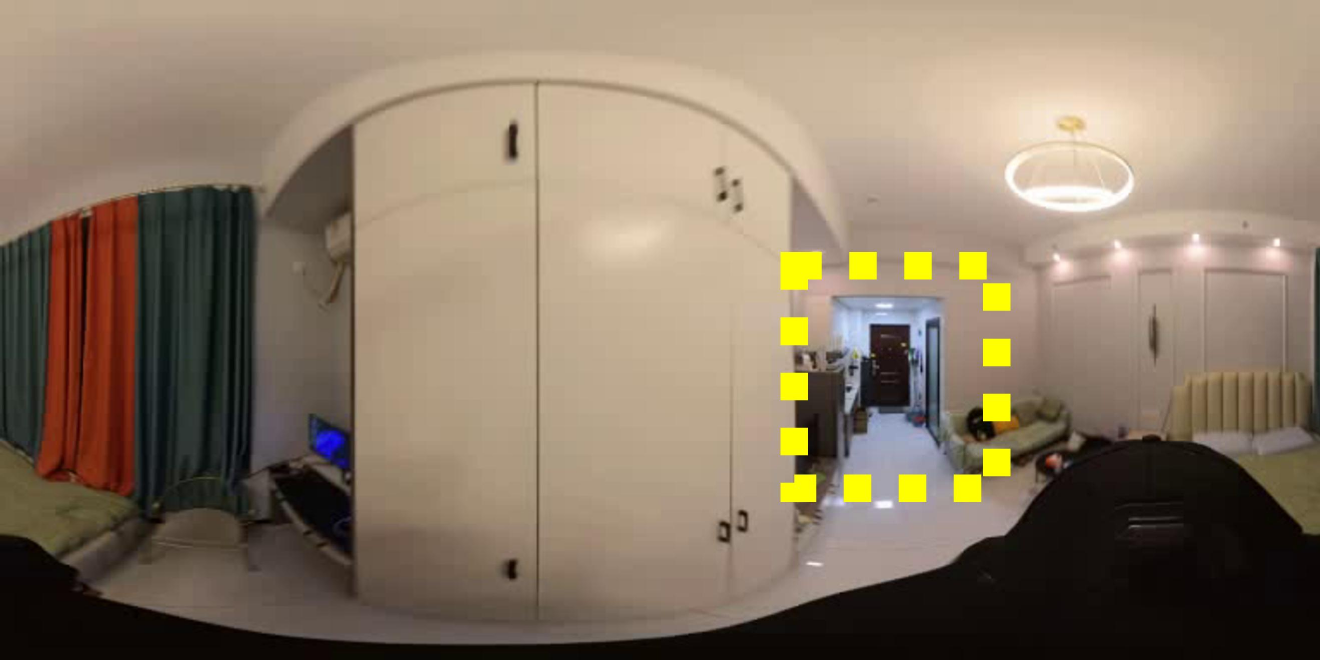}}&
\raisebox{-0.5\height}{\includegraphics[width=0.185\textwidth]{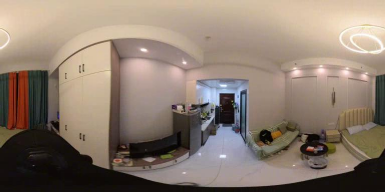}} &
\raisebox{-0.5\height}{\includegraphics[width=0.185\textwidth]{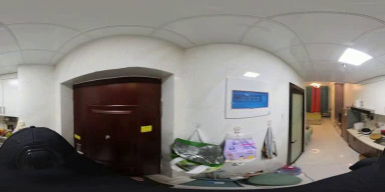}} &
\raisebox{-0.5\height}{\includegraphics[width=0.185\textwidth]{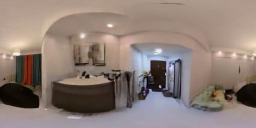}}&
\raisebox{-0.5\height}{\includegraphics[width=0.185\textwidth]{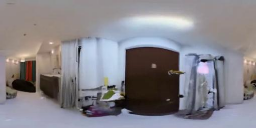}}\\
&$s_0$&$s_{T/2}$ GT &$s_T$ GT& $s_{T/2}$ Pred & $s_T$ Pred
\end{tabular}
\caption{Representative failure cases: anchor confusion, occlusion-induced detail loss, and long-range spatial compression.}
\label{fig:failure}
\vskip-4ex
\end{figure}

\section{Conclusion}
In this work, we address emulative simulation, where a virtual agent mentally explores the environment and reasons along the way. We introduce WanderDream, which consists of WanderDream-Gen for training and evaluating world models in imagining paths to target situations, and WanderDream-QA for assessing MLLMs in reasoning along imagined trajectories. Experiments show that imagination facilitates reasoning, confirming the correlation between world-model imagination and MLLM reasoning, and on our curated real-world test set the dataset exhibits sim-to-real transferability under occlusions and nonideal trajectories, though broader generality remains to be verified on larger and more diverse real-world data. We hope this work inspires further progress in imagination for exploring inaccessible real-world areas, and in enabling models to interpret human commands in virtual worlds, visualize imagined trajectories, and reason accordingly.

\section*{Acknowledgments}
We thank Weicheng Dai, Jingqi Zhang, and Zirui Wang for joining the human annotation and evaluation. This work was supported in part by the Ministry of Science, Research and the Arts of Baden-W\"urttemberg (MWK) through the Cooperative Graduate School Accessibility through AI-based Assistive Technology (KATE) under Grant BW6-03, in part by funding from the pilot program Core-Informatics of the Helmholtz Association (HGF), in part by Karlsruhe House of Young Scientists (KHYS), and in part by the Helmholtz Association Initiative and Networking Fund on the HAICORE@KIT and HOREKA@KIT partition. 
This work was supported in part by the National Natural Science Foundation of China (Grant No. 62473139, Grant No. 62503166), in part by the Hunan Provincial Research and Development Project (Grant No. 2025QK3019, Grant No. 2026QK3018), in part by the State Key Laboratory of Autonomous Intelligent Unmanned Systems (the opening project number ZZKF2025-2-10), in part by the Yuelushan Industrial Innovation Center, and in part by the Deutsche Forschungsgemeinschaft (DFG, German Research Foundation) - SFB 1574 - 471687386.
This research was partially funded by the Ministry of Education and Science of Bulgaria (support for INSAIT, part of the Bulgarian National Roadmap for Research Infrastructure).

%
%
\bibliographystyle{splncs04}
\bibliography{main}
\clearpage
\setcounter{page}{1}
\appendix


\section{Details of Data Generation}
The data generation process of WanderDream includes video generation for WanderDream-Gen and QA generation for WanderDream-QA.
\subsection{Video Generation for WanderDream-Gen}
To construct robotic situated imaginations in HM3D~\cite{ramakrishnan2021habitat}, we frame each episode as an object navigation task and treat the target object as the situational anchor. 
The complete generation procedure is outlined in Algorithm~\ref{alg:pano_hm3d}.

To construct human situated imaginations in ScanNet++~\cite{yeshwanth2023scannet++}, we first sample situations (\texttt{interacting}, \texttt{sitting}, \texttt{standing}) following Situat3DChange~\cite{liusituat3dchange}. We then sample a random start location at a distance between $1.5$m and $3$m from the target situation with a random orientation. If the direct path contains no non-traversable obstacles, the agent imagines the shortest path~\cite{Epstein2017CognitiveMap} and we directly interpolate between the start and target positions. If non-traversable obstacles are present, a 3D Probabilistic Roadmap (PRM) is used to plan the shortest path, as detailed in Algorithm~\ref{alg:flythrough}. 

\begin{figure*}
    \centering
    \includegraphics[width=\linewidth]{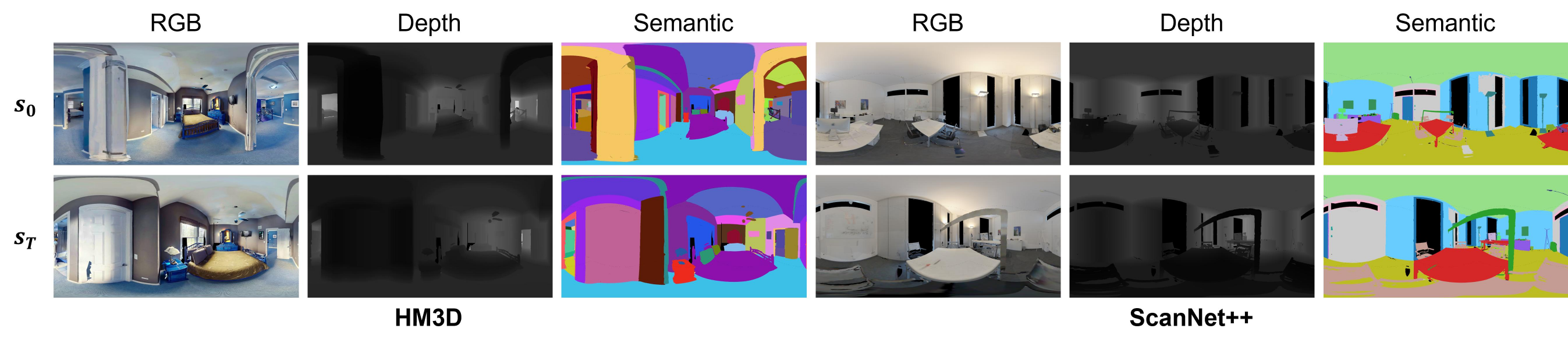}
    \caption{Visual modalities.}
    \label{fig:modalities}
\end{figure*}

The videos are generated as six directional views and then stitched into panoramic videos using the toolbox~\cite{zhang2014panocontext}. Each panoramic video is paired with corresponding depth and semantic modalities, as illustrated in Fig.~\ref{fig:modalities}, and will be released together with the associated camera poses.

\subsection{QA Generation for WanderDream-QA}
\begin{figure*}
    \centering
    \includegraphics[width=\linewidth]{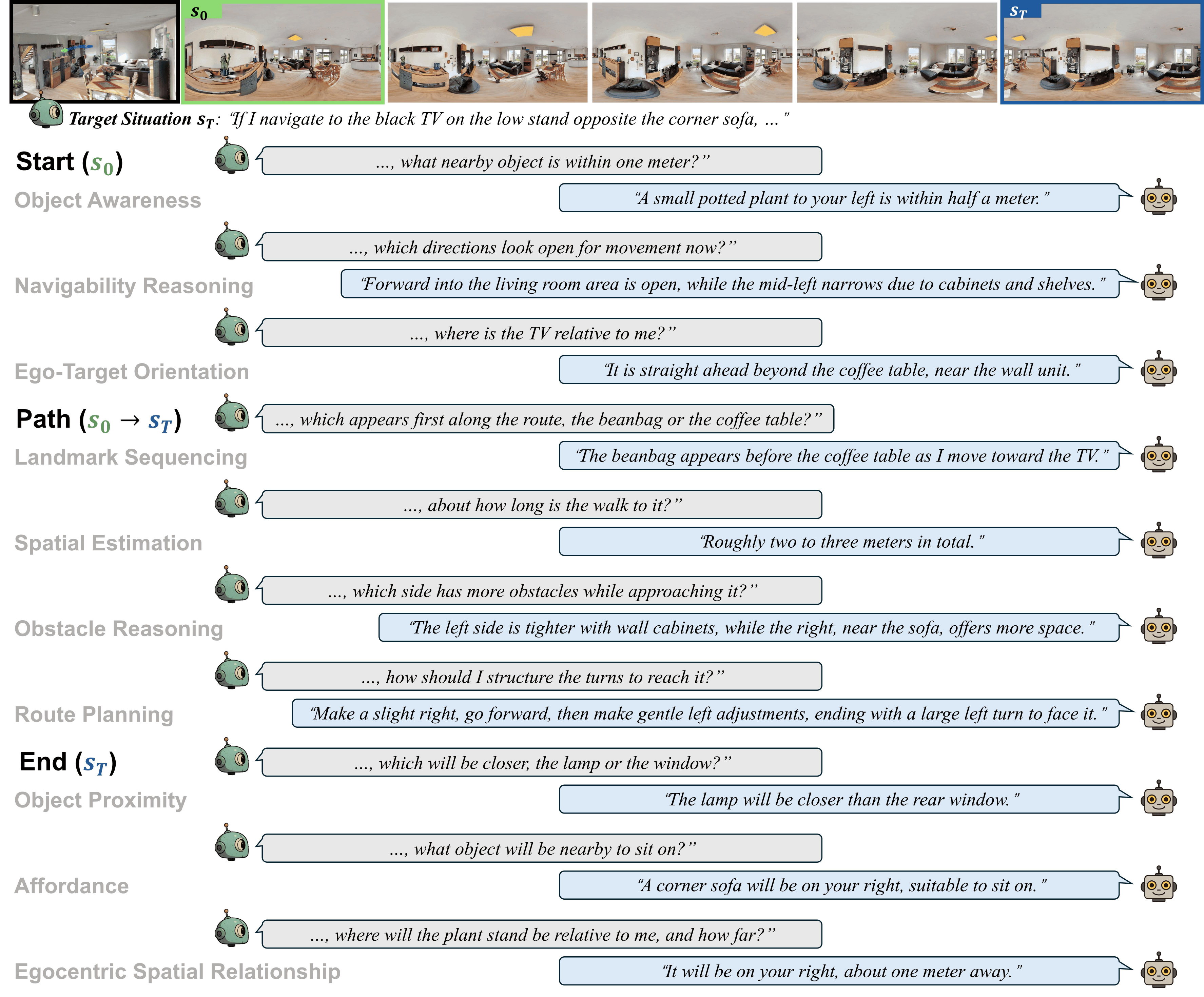}
    \caption{QA types from the robot perspective in HM3D object navigation.}
    \label{fig:qa_types_robot}
\end{figure*}

\begin{figure*}
    \centering
    \includegraphics[width=\linewidth]{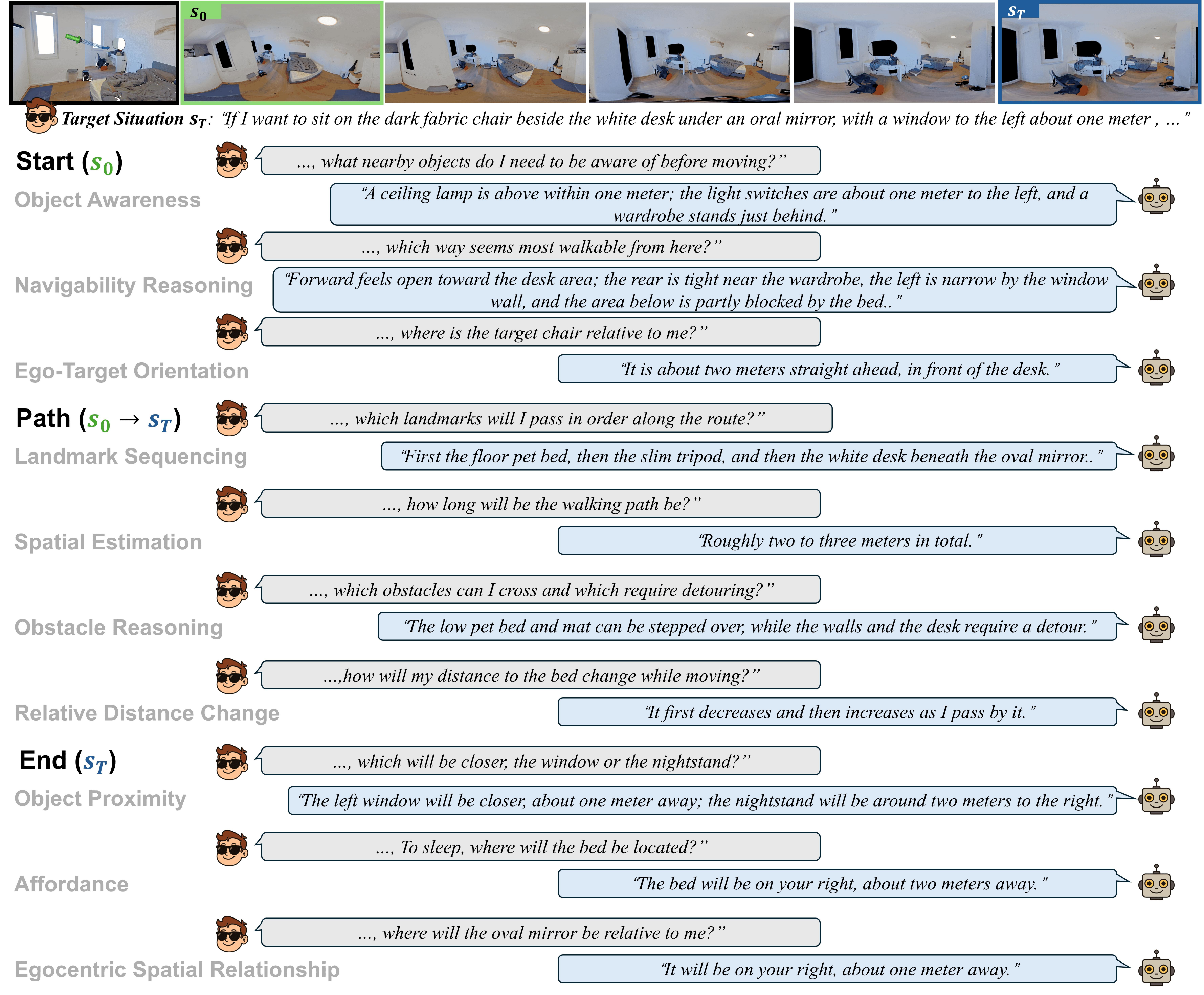}
    \caption{QA types from the human perspective in ScanNet++ scenes.}
    \label{fig:qa_types_human}
\end{figure*}

QA generation is conducted using GPT-5~\cite{openai2025gpt5} based on ground truth annotations. The prompt used for generating QA from the robot’s perspective is shown in Fig.~\ref{fig:robot_qa_gen_hm3d}, while the prompt from the human perspective is shown in Fig.~\ref{fig:human_qa_gen_prompt}. Representative generated QA examples are illustrated in Fig.~\ref{fig:qa_types_robot} and Fig.~\ref{fig:qa_types_human}.

\subsection{Data Statistics}
\begin{figure}
    \centering
    \includegraphics[width=0.7\linewidth]{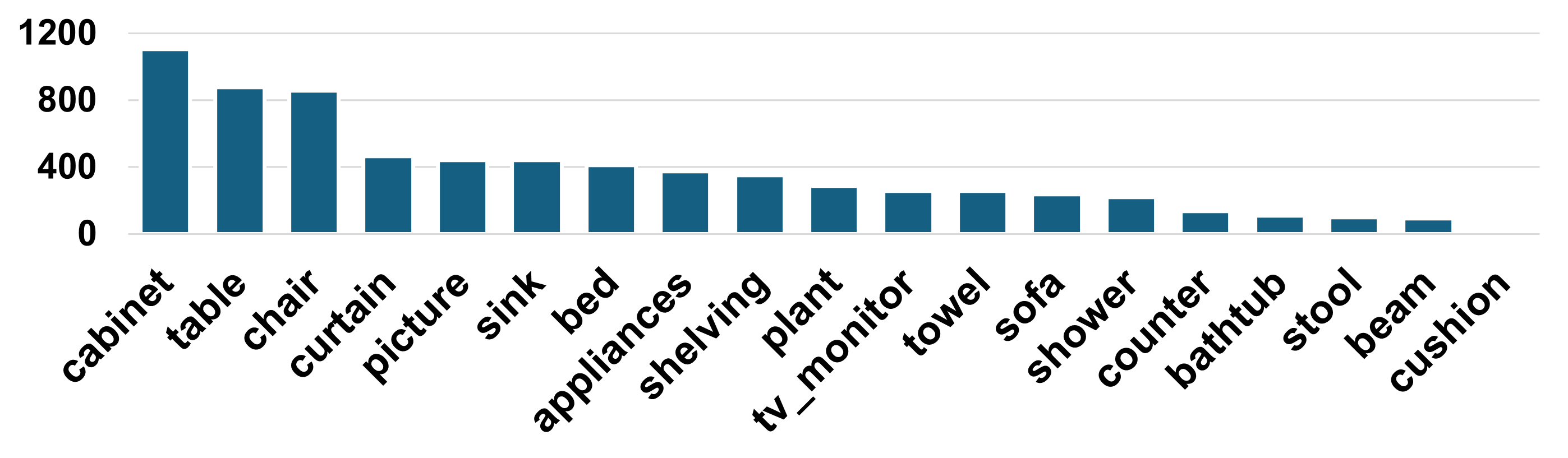}
    \caption{Target object categories for object navigation in HM3D.}
    \label{fig:landmarks}
\end{figure}

\begin{figure}
    \centering
    \includegraphics[width=0.7\linewidth]{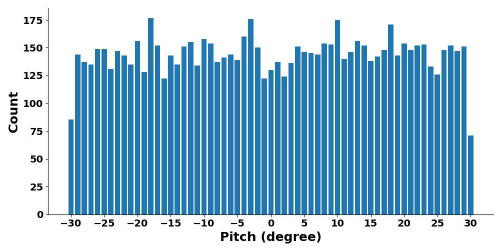}
    \caption{Histogram of the pitch used to mimic the human start state in ScanNet++.}
    \label{fig:histogram_pitch_start_state}
\end{figure}

Additional dataset statistics are provided. For WanderDream-Gen, we use $19$ object categories as target objects for object navigation and, consequently, as anchor objects for robot situation sampling. Their counts are shown in Fig.~\ref{fig:landmarks}. The counts of the three human situation types are reported in Tab.~\ref{tab:situ_num}. The histogram of the randomly sampled pitch angles at the start of human situations is shown in Fig.~\ref{fig:histogram_pitch_start_state}.

\begin{figure}[t]
    \centering
    \begin{subfigure}[b]{0.3\textwidth}
        \centering
        \includegraphics[width=\textwidth]{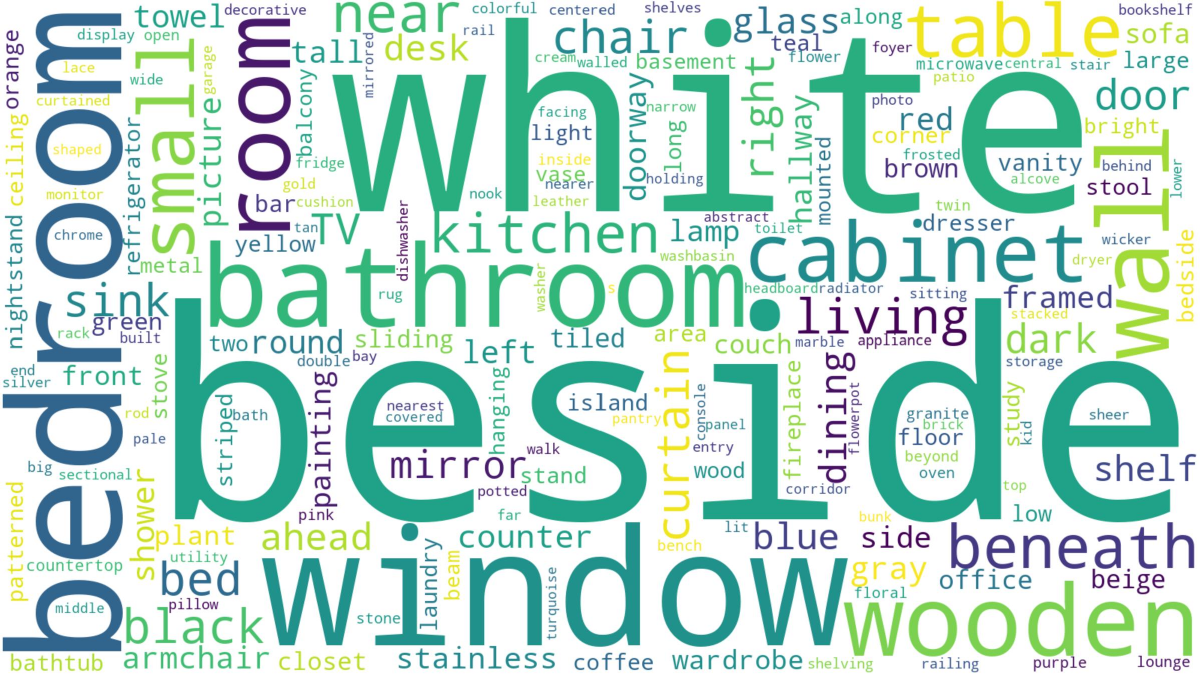}
        \caption{Robot situation.}
        \label{fig:sub1}
    \end{subfigure}
    \begin{subfigure}[b]{0.3\textwidth}
        \centering
        \includegraphics[width=\textwidth]{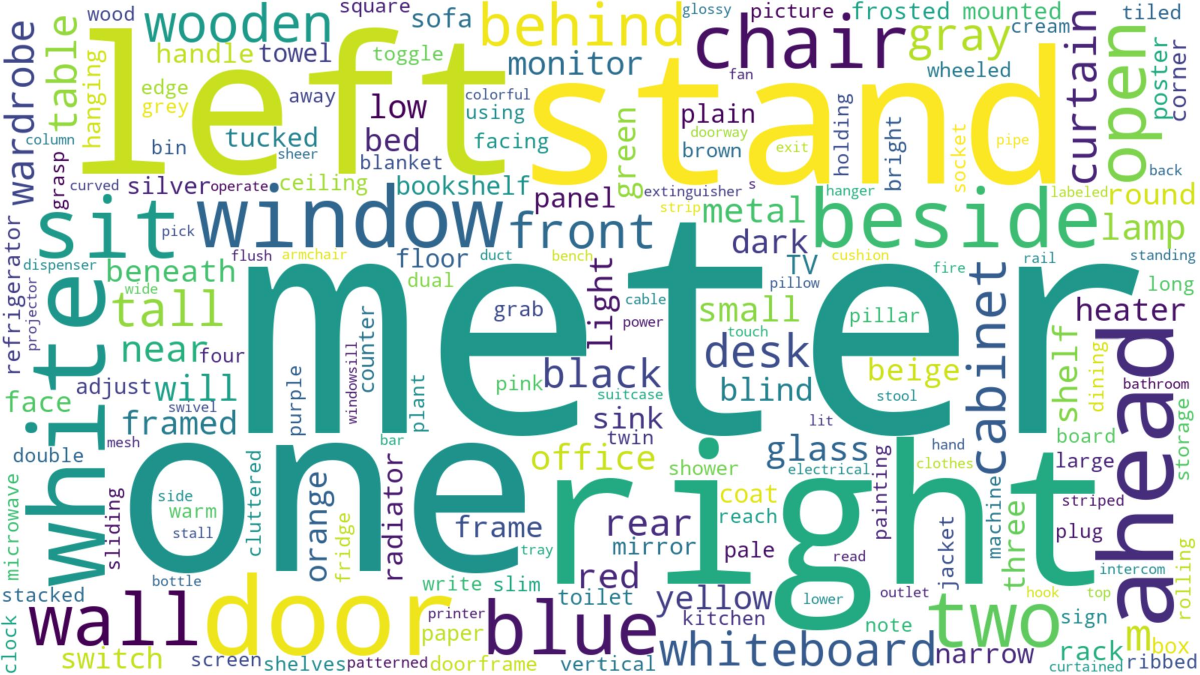}
        \caption{Human situation.}
        \label{fig:sub2}
    \end{subfigure}
    \begin{subfigure}[b]{0.3\textwidth}
        \centering
        \includegraphics[width=\textwidth]{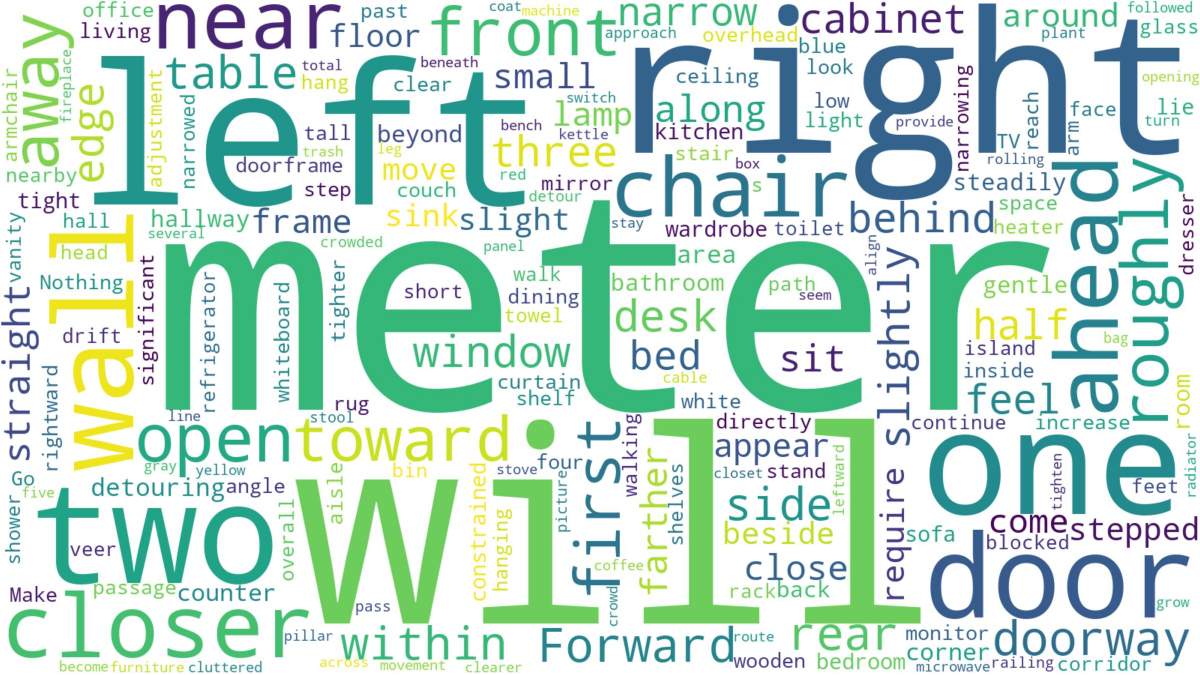}
        \caption{Overall answer.}
        \label{fig:sub3}
    \end{subfigure}

    \caption{Word clouds.}
    \label{fig:word_cloud}
\end{figure}

\begin{figure}[t]
    \centering
    \includegraphics[width=0.7\linewidth]{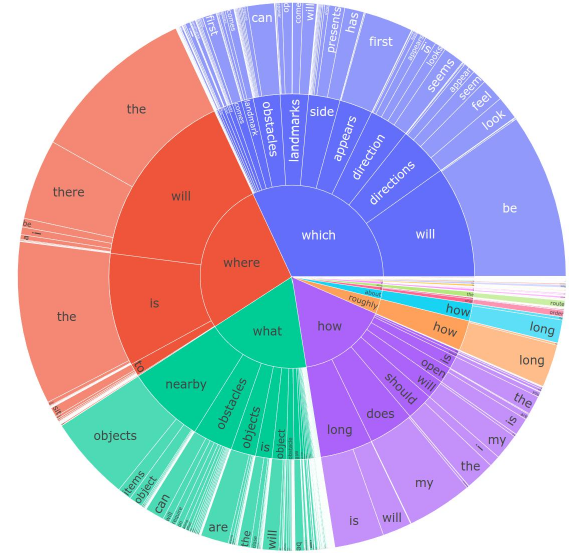}
    \caption{Hierarchical distribution of questions.}
    \label{fig:sunburst}
\end{figure}

For WanderDream-QA, Fig.~\ref{fig:word_cloud} shows the language diversity of the descriptive situations and answers, while Fig.~\ref{fig:sunburst} shows the hierarchical distribution of the questions.

\section{Experiment Details}
Fig.~\ref{fig:video_gen_prompt} shows the prompt template of the video generation pipeline. As mentioned in the main text, existing video generation models are mainly trained to generate foreground events, whereas our task requires the camera to move within the scene. Therefore, we either apply prompt extension, using an MLLM to locate the target object from the start state and to extend the original prompt with a description of the camera motion, or we fine-tune the models on WanderDream-Gen. Each video generation model has its own prompt extension script, and the script for Wan is shown in Fig.~\ref{fig:rewrite_video_prompt}. To fine-tune the video generation models, we follow the original settings, as summarized in Tab.~\ref{tab:training_config}.

Once the imagined exploration video towards the target situation is generated, we use an MLLM for reasoning, applying the template in Fig.~\ref{fig:eval_prompt_templates_different_input_settings} to predict the answer from this video.

\section{Evaluation}
Apart from the video generation scores (FVD, End-FID, S-SSIM, LPIPS) that assess generation quality, we follow the LLM-as-a-judge protocol~\cite{li2025generation} to evaluate whether the MLLM can reason about space based on the generated videos. Following~\cite{liusituat3dchange, ying2025mmwalk}, we use GPT-4o-mini with temperature set to $0$ to eliminate randomness. The scoring prompt is shown in Fig.~\ref{fig:score_prompt_qa}.

Note that we follow the established ground truth generation and evaluation strategy~\cite{ying2025mmwalk, linghu2024multimodal_situated_3d}. Since our long-form ground-truth answers are generated by GPT from ground-truth annotations, we evaluate predictions with MLLM-based scoring instead of lexical overlap metrics (\textit{e.g.}, BLEU and ROUGE). Lexical metrics can over-reward baselines that mimic the GPT-generated phrasing, even when the predicted content is not semantically correct. The widely studied self-preference bias of using the same MLLM as both a baseline and scorer~\cite{wataoka2024self, zheng2023judging} does not apply here, as the candidate answers being scored are generated by other models under evaluation, not by GPT itself. 

In Tab.~\ref{tab:llmjudge_ablation}, we conduct ablations over different LLM judges and prompt templates. To examine whether verbosity bias~\cite{marioriyad2025silent} influenced the evaluation results, we explicitly included the instruction shown in Fig.~\ref{fig:verbosity_avoid} to mitigate this bias. 
All configurations show strong alignment with human judgments (Spearman $\rho > 0.8$), while our setting achieves the highest correlation.

\section{Quantitative Analysis}
We design a comprehensive QA taxonomy to investigate when imagination facilitates situated reasoning. To analyze its role across QA classes with respect to anchor localization, we report anchor visibility at the start state in Tab.~\ref{tab:visible_statistics} and per-class performance with ground-truth frames in Tab.~\ref{tab:gt}. The anchor visibility in Tab.~\ref{tab:visible_statistics} also quantifies the proportion of questions that cannot be answered from the start state alone, i.e., questions that would be ill-posed without imagination. Anchor objects are visible in nearly all ScanNet++ cases at the start state, but in $28.3\%$ of HM3D validation cases they are not. Across all end-state $s_T$ classes, high-quality imagination remains necessary regardless of anchor visibility. Under severe occlusion, imagination improves Spatial Estimation and Route Planning in HM3D. In ScanNet++, where anchor visibility is high, path-related questions can often be resolved from the start state alone, and directional cues may already be implied by the question text for Landmark Sequencing and Obstacle Reasoning. In such cases, imagination may even degrade reasoning performance by introducing orientation distractions. As MLLMs improve, we expect the benefit of imagination to become more pronounced.

\section{Qualitative Analysis}
Fig.~\ref{fig:qualitative_results_hm3d} presents a set of qualitative results in HM3D. CogVideoX with prompt extension fails to move the camera and instead generates a person approaching the target object (a sink). Wan2.2 with prompt extension moves toward the sink correctly but does not preserve the panoramic structure, collapsing to a front‐view perspective. HunyuanVideo produces blurry videos, whereas the other models even reconstruct the occluded oven in the first frame correctly. Fig.~\ref{fig:qualitative_results_scannetpp} shows ScanNet++ results with similar trends. The fine-tuned models maintain the correct layout in the final state, whereas CogVideoX with prompt extension again introduces a person to carry out the action.


\section{Limitations and Future Work}
At present, the model performs imagination solely based on the current egocentric observation. In future work, we plan to incorporate previous views~\cite{zhou2025learning} and longer-term memory~\cite{yu2025vismem, yang20253d}. We will also release more visual modalities, including depth, semantics, and camera poses. Although they are not used in this paper, these modalities are expected to further strengthen the emulative simulation, leading to more stable and better-controlled video generation. 
Furthermore, WanderDream demonstrates strong sim-to-real performance in imagining situations with a panorama mounted on the agent, even when parts of the scene are occluded by the agent. These occlusion cases are not included in the training set. In future work, we plan to collect more real-world data with cameras mounted on different types of agents that have varying mobility characteristics, enabling us to scale emulative simulation to broader cases and scenarios.

\section{Societal Impacts}
WanderDream enables emulative simulation by allowing models to \textit{mentally walk} through an environment, generating a plausible visual trajectory toward an envisioned situation, and performing situated reasoning along this path. In addition to our core motivation of reducing the physical limitations of robots and the psychological burden on blind and visually impaired people when they try to understand cluttered scenes through an imagined situation, WanderDream provides a general mechanism for anticipating and analyzing future states of the world. This capability can naturally be transferred to a broad spectrum of applications, \textit{e.g.}, autonomous driving~\cite{wang2024stag-1}, pedestrian assistance~\cite{hassan2025gem}, virtual real estate exploration~\cite{ccelen2025housetour}, and other interactive decision-making scenarios that require anticipating how situations evolve over space and time.

\begin{table}[]
    \centering
    \caption{Situations on ScanNet++ from human perspectives.}
    \begin{tabular}{c|c|c|c}
    \toprule
         & interacting& sitting& standing \\
    \midrule
         Train&2772&1903&3599\\
         Validation&181&76&221\\
    \midrule
         Total&2953&1979&3820\\
    \bottomrule
    \end{tabular}
    \label{tab:situ_num}
\end{table}

\begin{table}[b!]
\centering
\setlength{\tabcolsep}{3pt}
\caption{Statistics of anchor object visibility at the start state $s_o$.}
\begin{tabular}{@{}c|c|c|c|c@{}}
\toprule
Subset & Split & Samples & Visible & Percentage (\%) \\
\midrule
\multirow{2}{*}{HM3D} & Train & 5632 & 3921 & 69.6 \\
                       & Val   & 1404 & 1006 & 71.7 \\
\midrule
\multirow{2}{*}{ScanNet++} & Train & 8274 & 7892 & 95.4 \\
                            & Val   & 478  & 453  & 94.8 \\
\bottomrule
\end{tabular}

\label{tab:visible_statistics}
\end{table}

\clearpage
\begin{algorithm}[t]
\caption{Panoramic-episode generation in HM3D object navigation}
\label{alg:pano_hm3d}
\KwIn{simulator $\mathsf{sim}$, target instance $i$}
\KwOut{six-view RGB / depth / semantic videos and pose list $\mathcal{P}$}

$\mathcal{G} \leftarrow \textsc{SampleGoalPoints}(\mathsf{sim}, i)$\;
$s_0 \leftarrow \textsc{RandomStart}(\mathsf{sim}, \mathcal{G})$\;
$\mathsf{sp} \leftarrow \textsc{MultiGoalShortestPath}(\mathsf{sim}, s_0, \mathcal{G})$\;

init 6-view RGB/depth/semantic writers\;
set agent at $s_0$ facing next waypoint\;

$\mathcal{P} \leftarrow \emptyset$\;

\ForEach{waypoint $g$ in $\mathsf{sp.points}$}{
  \While{agent not near $g$ and step $<$ max steps}{
    $a \leftarrow \textsc{NextGreedyStep}(\mathsf{sim}, g)$\;
    apply $a$ in $\mathsf{sim}$\;
    render six-view RGB / depth / semantic\;
    save all views, append pose to $\mathcal{P}$\;
  }
}
\textsc{OrientToTarget}$(\mathsf{sim}, i.center)$
\end{algorithm}

\begin{algorithm}[t]
\caption{Fly-through generation with straight-line fallback in ScanNet++}
\label{alg:flythrough}
\KwIn{scan, standable mask, situation $(\mathbf{p}_T,\mathbf{d}_T)$}
\KwOut{trajectory, cubemap videos, \texttt{path.json}}

sample start eye $\mathbf{p}_0$ in standable region\;

\eIf{\textsc{SegClear}($\mathbf{p}_0,\mathbf{p}_T$)}{
  $\{\mathbf{p}_t\}_{t=0}^{F-1} \leftarrow \textsc{LinInterp}(\mathbf{p}_0,\mathbf{p}_T,F)$\;
  choose start dir $\mathbf{d}_0$ from $\mathbf{d}_T$\;
  $\{\mathbf{d}_t\}_{t=0}^{F-1} \leftarrow \textsc{DirInterp}(\mathbf{d}_0,\mathbf{d}_T)$\;
}{
  $(\mathcal{P},L) \leftarrow \textsc{PlanPRM}(\mathbf{p}_0,\mathbf{p}_T)$\;
  $(\mathcal{P},ok) \leftarrow \textsc{CapCheck}(\mathcal{P})$\;
  \If{not $ok$}{abort and log failure\;}
  $(\{\mathbf{p}_t\},\{u_t\},L_{\text{tot}}) \leftarrow \textsc{Resample}(\mathcal{P},F)$\;
  $\{\mathbf{p}_t\} \leftarrow \textsc{MonoZ}(\{\mathbf{p}_t\})$\;
  choose start dir $\mathbf{d}_0$ from $\mathbf{d}_T$\;
  $\{\mathbf{d}_t\} \leftarrow \textsc{DirInterp}(\mathbf{d}_0,\mathbf{d}_T;\{u_t\})$\;
}

\For{$t \gets 0$ \KwTo $F-1$}{
  nudge $\mathbf{p}_t$ away from mesh if needed\;
}
\If{not \textsc{PathCheck}($\{\mathbf{p}_t\}$)}{abort and log failure\;}

save $\{\mathbf{p}_t,\mathbf{d}_t\}$ to \texttt{path.json}\;
render six-view RGB / depth / semantic\;
save all views, append pose to $\mathcal{P}$\;
\end{algorithm}


\begin{figure*}[]
\fontsize{7.5}{9}\selectfont
\begin{tcolorbox}
\begin{minipage}{\linewidth}

\textcolor{blue}{system\_prompt} = (``%
Generate 10 QA pairs based on 15 images, including:\\
-- start position and end position, each with six cube faces (up, left, front, right, rear, down) containing a Set of Marks;\\
-- three route images.\\[0.3em]
Also include a summary JSON describing:\\
-- target object;\\
-- the trajectory;\\
-- the Set of Marks information at both start and end positions.\\[0.3em]
Taking the provided JSON as an example, only output the situation and QA pairs in a JSON array, without any additional text.\\[0.3em]
The format of the output JSON file is:\\
\{\texttt{"situation": "", "qa": []}\}.\\[0.3em]
Each QA must follow this format:\\
\{\texttt{"phase": "", "question": "", "answer": "", "type": ""}\}.\\[0.3em]
The overall \texttt{"situation"} field should be a short description of the target object using ``If I navigate to \dots'':\\
-- if visible in the start view, distinguish it from distractors by color or spatial relation, and keep it very brief;\\
-- if not visible, mention the room type it is located in, in addition to distinguishing features (e.g., ``the TV on the cabinet in the living room'').\\
Treat the situation as a half-sentence condition for the QA pairs, make it brief and avoid overlapping its information with the question or answer.\\[0.3em]
Each question and answer must be within 30 words. Do not list things; only describe.\\
If distractors of the mentioned objects exist, use concise distinguishing features to specify the object.\\
At the start phase, reason only from the start images; at the end phase, use future tense (``will be'', ``will have'').\\
There must be exactly 10 QA pairs, strictly covering 10 distinct question types.\\[0.5em]
Question type descriptions:\\[0.25em]
Start phase (3):\\
1. Object Awareness: describe objects in natural language by approximate distance (e.g., ``within one meter'', ``far away'').\\
2. Navigability Reasoning: analyze pathable areas and spatial openness in all directions.\\
3. Egocentric Direction: describe the rough direction of the target relative to the observer.\\[0.25em]
Path phase (4):\\
4. Landmark Sequencing: compare the ordered sequence of landmarks encountered along the route.\\
5. Spatial Estimation: roughly estimate path length using meters but without precise numbers.\\
6. Obstacle Reasoning: describe obstacles on one side of the path.\\
7. Route Planning: describe route structure or turns.\\[0.25em]
End phase (3):\\
8. Object Proximity: compare which of two objects is closer (e.g., a couch about four meters away). Do not compare the target object here.\\
9. Affordance: query the presence or location of a functional object (e.g., to sit on, to heat water, to light the room, or food to eat). Use an infinitive in the question, not an intention phrase.\\
10. Egocentric Spatial Relation: determine the egocentric direction of an object (left/right/front/rear) with rough distance (meter).'')

\end{minipage}
\end{tcolorbox}

\caption{Prompt for LLM-based QA generation in WanderDream-QA from the robot perspective using HM3D data.}
\label{fig:robot_qa_gen_hm3d}
\end{figure*}

\begin{figure*}[!h]
\fontsize{6.8}{7.4}\selectfont
\begin{tcolorbox}
\begin{minipage}{\linewidth}

\textcolor{blue}{situ\_prompts} = \{\\
\quad \texttt{"sitting"}:\ ``Describe the seat using very concise distinguishing features compared to other seats, and mention one nearby landmark with its direction and approximate distance (integer meter).'',\\
\quad \texttt{"interacting"}:\ ``Use a verb other than `interact'; describe the object using very concise distinguishing features, and mention another nearby landmark in a different direction (not in front) with its approximate distance (integer meter).'',\\
\quad \texttt{"standing"}:\ ``Use two landmarks with distinguishing features in two directions, including their distances (integer meter).''\\
\}\par\vspace{0.5em}

\textcolor{blue}{system\_prompt} = (``%
Generate 10 QA pairs based on 15 images, including:\\
-- Start position and end position, each with six cube faces (up, left, front, right, rear, down) containing a Set of Marks.\\
-- Three route images.\\[0.3em]
Also include a summary JSON describing:\\
-- brief target situation;\\
-- position change and video path length;\\
-- the Set of Marks information at both start and end positions.\\[0.3em]
Taking the provided JSON as an example, only output the situation and QA pairs in a JSON array, without any additional text.\\[0.3em]
The format of the output JSON file is:\\
\{\texttt{"situation": "", "qa": []}\}.\\[0.3em]
Each QA must follow this format:\\
\{\texttt{"phase": "", "question": "", "answer": "", "type": ""}\}.\\[0.3em]
The overall \texttt{"situation"} field should describe the end state based on the provided short situation with ``If I want to''. Note it is not the start state; it is imagined.\\
\{\textcolor{blue}{situ\_prompts[situ\_cls]}\}\\[0.3em]
Treat the situation as a half-sentence condition for the QA pairs, make it brief, and avoid overlapping its information with the question or answer.\\[0.3em]
Each question and answer must be within 30 words. Do not list items; only describe.\\
If distractors of the mentioned objects exist, use concise distinguishing features (e.g., pattern, location, color) to specify the object; do not directly compare.\\
At the start phase, reason only from the start images; at the end phase, use future tense (``will be'', ``will have'').\\
There must be exactly 10 QA pairs, strictly covering 10 distinct question types.\\
Note that the rooms are relatively small, so over two meters means ``far''.\\[0.5em]
Question type descriptions:\\[0.25em]
Start phase (3):\\
1. Object Awareness: describe objects around the start position in natural language with approximate distance (e.g., ``within one meter'', ``far away'').\\
2. Navigability Reasoning: analyze pathable areas and spatial openness in all directions.\\
3. Egocentric Direction: describe the rough direction of the target relative to the observer.\\[0.25em]
Path phase (4):\\
4. Landmark Sequencing: compare the ordered sequence of landmarks encountered along the route within arm reach; do not summarize.\\
5. Spatial Estimation: roughly estimate path length using integer meters.\\
6. Obstacle Reasoning: summarize which obstacles on the path can be crossed or stepped over (e.g., trash bin) and which ones require detouring (e.g., wall). If nothing, answer ``nothing significant''.\\
7. Relative Distance Change: describe whether the observer moves closer, farther, or first closer then farther (pass by) to a specific object.\\[0.25em]
End phase (3):\\
8. Object Proximity: compare which of two objects is closer (e.g., couch about four meters away). Do not compare the target object here.\\
9. Affordance: query the presence or location of a functional object (e.g., to sit on, to heat water, or food to eat). Use an infinitive in the question, not an intention phrase. Do not ask about the object explicitly mentioned in the situation, but you may ask about another object of the same type or a distractor with similar function.\\
10. Egocentric Spatial Relation: determine the egocentric direction of an object (left/right/front/rear) with rough distance (meter).'')

\end{minipage}
\end{tcolorbox}

\caption{Prompt for LLM-based QA generation in WanderDream-QA from the human perspective using ScanNet++ data.}
\label{fig:human_qa_gen_prompt}
\end{figure*}

\clearpage

\begin{figure*}[]
\begin{tcolorbox}
\begin{minipage}{\linewidth}
``Starting from a panoramic view at the initial location, generate an egocentric
panoramic sequence that moves toward the target position: {\textcolor{blue}{situation\_cap}\}}''

\end{minipage}
\end{tcolorbox}

\caption{Prompt template for video generation.}
\label{fig:video_gen_prompt}
\end{figure*}
\begin{figure*}[!h]
\fontsize{7.5}{9}\selectfont
\begin{tcolorbox}
\begin{minipage}{\linewidth}

\textcolor{blue}{system\_prompt} = (``%
You are an expert in rewriting video description prompts. Your task is to rewrite the provided video description prompts based on the images given by users, emphasizing potential dynamic content.\\[0.3em]
The user's input language may include diverse descriptions, such as markdown format, instruction format, or be too long or too short. You need to extract the relevant information from the user’s input and associate it with the image content.\\[0.3em]
Your rewritten video description should retain the dynamic parts of the provided prompts, focusing on the main subject's actions. Emphasize and simplify the main subject of the image while retaining their movement. If the user only provides an action (e.g., `dancing'), supplement it reasonably based on the image content (e.g., `a girl is dancing').\\[0.3em]
If the user’s input prompt is too long, refine it to capture the essential action process. If the input is too short, add reasonable motion-related details based on the image content.\\[0.3em]
Retain and emphasize descriptions of camera movements, such as `the camera pans up', `the camera moves from left to right', or `the camera moves from right to left'. For example: `The camera captures two men fighting. They start lying on the ground, then the camera moves upward as they stand up. The camera shifts left, showing the man on the left holding a blue object while the man on the right tries to grab it, resulting in a fierce back-and-forth struggle.'\\[0.3em]
Focus on dynamic content in the video description and avoid adding static scene descriptions. If the user’s input already describes elements visible in the image, remove those static descriptions.\\[0.3em]
Limit the rewritten prompt to 100 words or less. Regardless of the input language, your output must be in English.\\[0.5em]
Examples of rewritten prompts:\\
-- The camera pulls back to show two foreign men walking up the stairs. The man on the left supports the man on the right with his right hand.\\
-- A black squirrel focuses on eating, occasionally looking around.\\
-- A man talks, his expression shifting from smiling to closing his eyes, reopening them, and finally smiling with closed eyes. His gestures are lively, making various hand motions while speaking.\\
-- A close-up of someone measuring with a ruler and pen, drawing a straight line on paper with a black marker in their right hand.\\
-- A model car moves on a wooden board, traveling from right to left across grass and wooden structures.\\
-- The camera moves left, then pushes forward to capture a person sitting on a breakwater.\\
-- A man speaks, his expressions and gestures changing with the conversation, while the overall scene remains constant.\\
-- The camera moves left, then pushes forward to capture a person sitting on a breakwater.\\
-- A woman wearing a pearl necklace looks to the right and speaks.\\
Output only the rewritten text without additional responses.'')

\end{minipage}
\end{tcolorbox}
\caption{Prompt extension for Wan.}
\label{fig:rewrite_video_prompt}
\end{figure*}

\begin{figure*}[!h]
\begin{tcolorbox}
\begin{minipage}{\linewidth}

  $s_0$:  
  ``This is my current panoramic view, and my target situation is \{\textcolor{blue}{situation\_cap}\}.\\
  Please answer the following question about \{\textcolor{blue}{state}\}: \{\textcolor{blue}{question}\}.''

  $(s_0,s_T)$ or $s_{\Delta 5}$:  
  ``This is a panoramic image sequence from my current view to the target situation \{\textcolor{blue}{situation\_cap}\}.\\
  Please answer the following question about \{\textcolor{blue}{state}\}: \{\textcolor{blue}{question}\}.''

  $s_v$:  
  ``This is a panoramic video from my current view to the target situation \{\textcolor{blue}{situation\_cap}\}.\\
  Please answer the following question about \{\textcolor{blue}{state}\}: \{\textcolor{blue}{question}\}.''  

\end{minipage}
\end{tcolorbox}

\caption{Prompt templates for different input settings.}
\label{fig:eval_prompt_templates_different_input_settings}
\end{figure*}

\begin{figure*}[!h]

\begin{tcolorbox}
\begin{minipage}{\linewidth}
\textcolor{blue}{system\_prompt} = (
``You are an intelligent evaluator designed to evaluate the correctness and similarity of generative outputs for question–answer pairs.
Your task is to compare the model prediction answer with the correct answer and determine if they match in meaning.

Scoring criteria:

5 = Perfect match or correct in meaning.

4 = Key information correct, minor flaws.

3 = Partially correct.

2 = Mostly wrong answer for key query, but some relevance.

1 = Completely wrong or nonsense sentences.

Your response must ONLY be the integer score (e.g., 4). DO NOT include any text or explanation.'')

\textcolor{blue}{user\_prompt} = (``

Question: {\{\textcolor{blue}{question}\}}

Correct Answer: \{\textcolor{blue}{gt\_answer}\}

Predicted Answer:\{\textcolor{blue}{pred\_answer}\}
                
Please provide a score from 1 to 5 based on how well the predicted answer matches the correct answer.'')

\end{minipage}
\end{tcolorbox}

\caption{Prompt for LLM-assisted scoring of WanderDream-QA.}
\label{fig:score_prompt_qa}

\end{figure*}
\begin{figure*}[]
\begin{tcolorbox}
\begin{minipage}{\linewidth}
``CRITICAL INSTRUCTION — Verbosity Bias Prevention:
        Answer length must NOT influence your score. A short, accurate answer is equally good as a long, 
        accurate answer. A long answer that adds irrelevant detail or padding is NOT better than a concise correct answer. Score ONLY on whether the key information in the correct answer is present and accurate in the predicted answer — never reward or penalize based on length alone.''

\end{minipage}
\end{tcolorbox}

\caption{Prompt instruction for explicitly avoiding verbosity bias.}
\label{fig:verbosity_avoid}
\end{figure*}
\begin{figure*}[t!]
\scriptsize
\setlength\tabcolsep{1pt}
\centering

\begin{tabular}{c c c c c c}
\rotatebox[origin=c]{90}{GT} &
\raisebox{-0.5\height}{\includegraphics[width=0.185\textwidth]{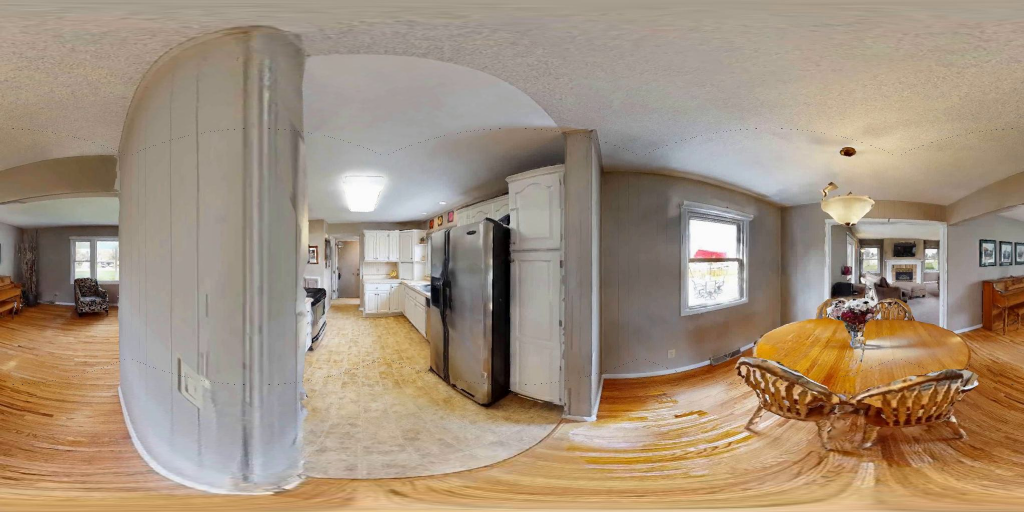}} &
\raisebox{-0.5\height}{\includegraphics[width=0.185\textwidth]{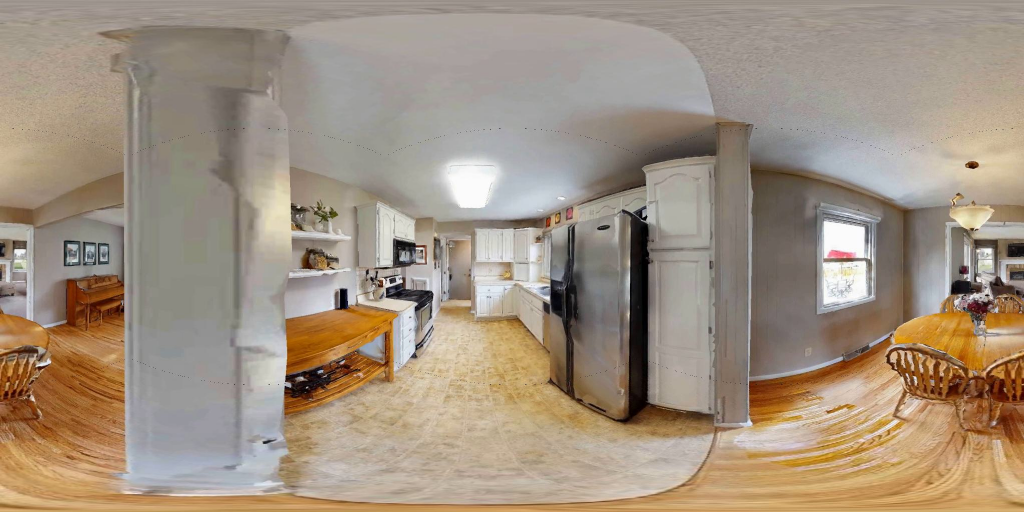}} &
\raisebox{-0.5\height}{\includegraphics[width=0.185\textwidth]{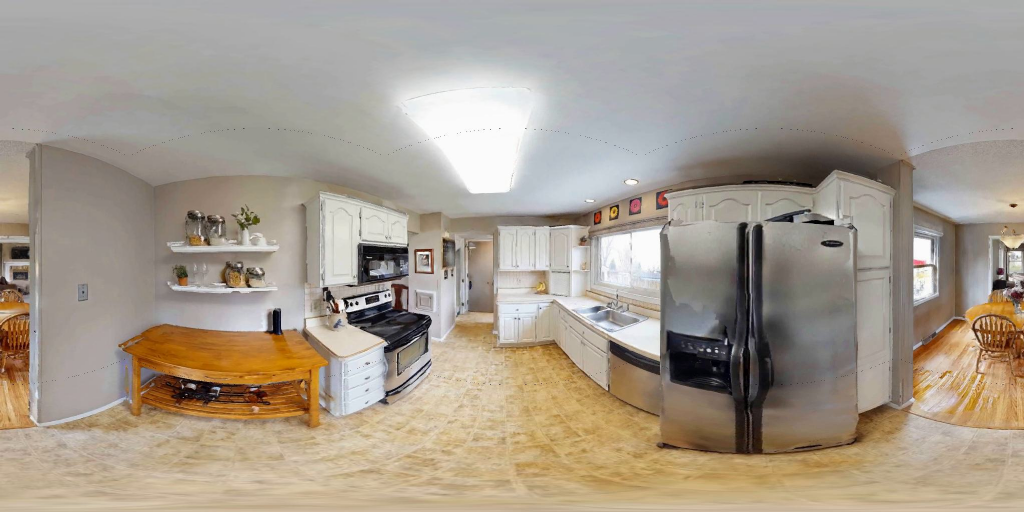}} &
\raisebox{-0.5\height}{\includegraphics[width=0.185\textwidth]{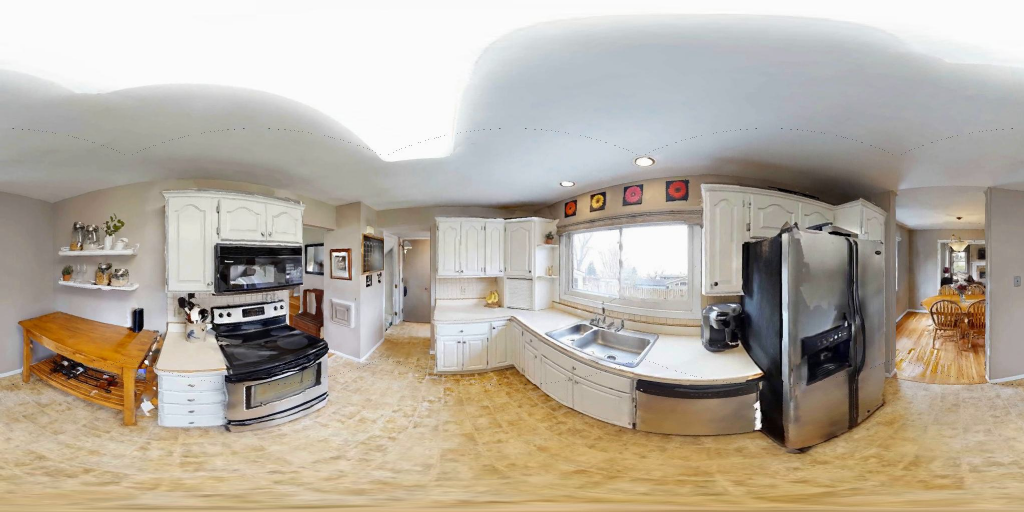}} &
\raisebox{-0.5\height}{\includegraphics[width=0.185\textwidth]{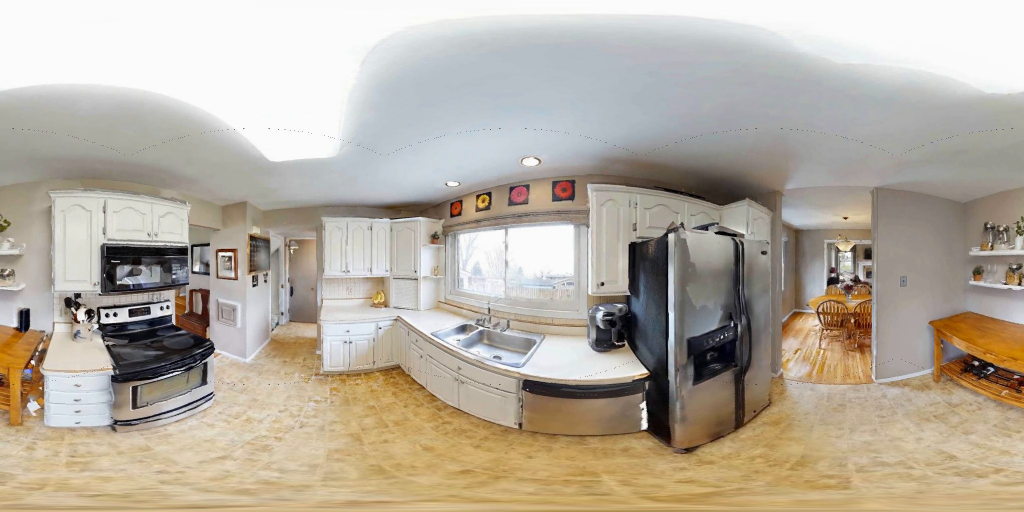}}\\
\rotatebox[origin=c]{90}{CogVideoX*} &
\raisebox{-0.5\height}{\includegraphics[width=0.185\textwidth]{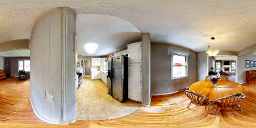}} &
\raisebox{-0.5\height}{\includegraphics[width=0.185\textwidth]{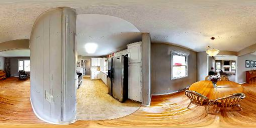}} &
\raisebox{-0.5\height}{\includegraphics[width=0.185\textwidth]{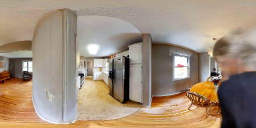}} &
\raisebox{-0.5\height}{\includegraphics[width=0.185\textwidth]{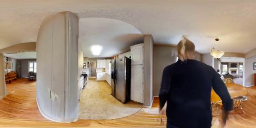}} &
\raisebox{-0.5\height}{\includegraphics[width=0.185\textwidth]{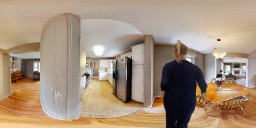}}\\
\rotatebox[origin=c]{90}{Wan2.2*} &
\raisebox{-0.5\height}{\includegraphics[width=0.185\textwidth]{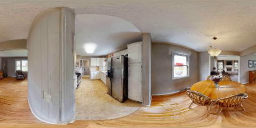}} &
\raisebox{-0.5\height}{\includegraphics[width=0.185\textwidth]{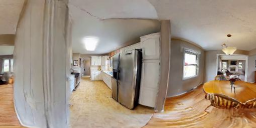}} &
\raisebox{-0.5\height}{\includegraphics[width=0.185\textwidth]{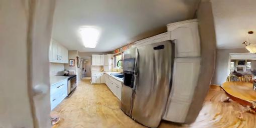}} &
\raisebox{-0.5\height}{\includegraphics[width=0.185\textwidth]{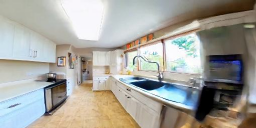}} &
\raisebox{-0.5\height}{\includegraphics[width=0.185\textwidth]{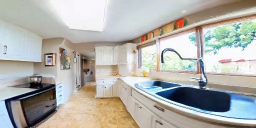}}\\
\rotatebox[origin=c]{90}{Hunyuan$^\dag$} &
\raisebox{-0.5\height}{\includegraphics[width=0.185\textwidth]{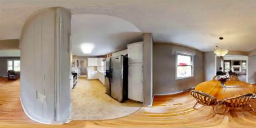}} &
\raisebox{-0.5\height}{\includegraphics[width=0.185\textwidth]{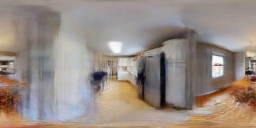}} &
\raisebox{-0.5\height}{\includegraphics[width=0.185\textwidth]{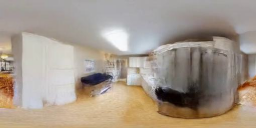}} &
\raisebox{-0.5\height}{\includegraphics[width=0.185\textwidth]{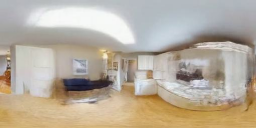}} &
\raisebox{-0.5\height}{\includegraphics[width=0.185\textwidth]{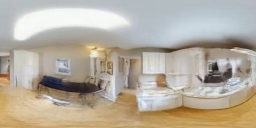}}\\
\rotatebox[origin=c]{90}{Wan2.1$^\dag$} &
\raisebox{-0.5\height}{\includegraphics[width=0.185\textwidth]{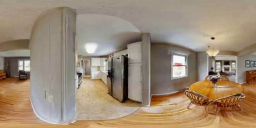}} &
\raisebox{-0.5\height}{\includegraphics[width=0.185\textwidth]{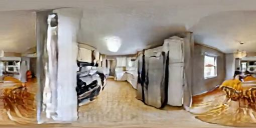}} &
\raisebox{-0.5\height}{\includegraphics[width=0.185\textwidth]{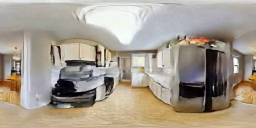}} &
\raisebox{-0.5\height}{\includegraphics[width=0.185\textwidth]{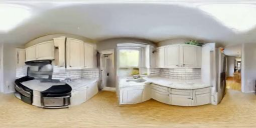}} &
\raisebox{-0.5\height}{\includegraphics[width=0.185\textwidth]{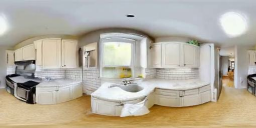}}\\
\rotatebox[origin=c]{90}{Wan2.2$^\dag$} &
\raisebox{-0.5\height}{\includegraphics[width=0.185\textwidth]{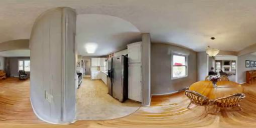}} &
\raisebox{-0.5\height}{\includegraphics[width=0.185\textwidth]{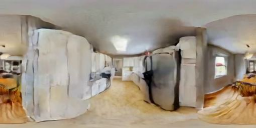}} &
\raisebox{-0.5\height}{\includegraphics[width=0.185\textwidth]{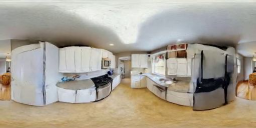}} &
\raisebox{-0.5\height}{\includegraphics[width=0.185\textwidth]{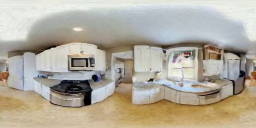}} &
\raisebox{-0.5\height}{\includegraphics[width=0.185\textwidth]{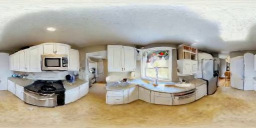}}\\
\rotatebox[origin=c]{90}{CogVideoX$^\dag$} &
\raisebox{-0.5\height}{\includegraphics[width=0.185\textwidth]{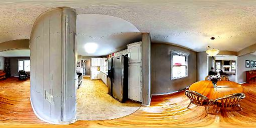}} &
\raisebox{-0.5\height}{\includegraphics[width=0.185\textwidth]{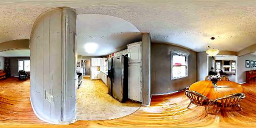}} &
\raisebox{-0.5\height}{\includegraphics[width=0.185\textwidth]{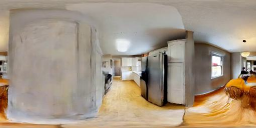}} &
\raisebox{-0.5\height}{\includegraphics[width=0.185\textwidth]{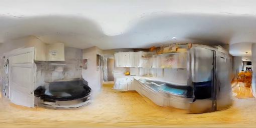}} &
\raisebox{-0.5\height}{\includegraphics[width=0.185\textwidth]{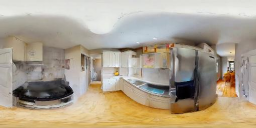}}\\
\end{tabular}
\caption{Qualitative results in HM3D for the situation \textit{``If I navigate to the sink beneath the wide window with flower pictures in the kitchen.''}. * denotes prompt extension, while $\dag$ denotes fine-tuning.}
\label{fig:qualitative_results_hm3d}
\end{figure*}

\begin{figure*}[t!]
\scriptsize
\setlength\tabcolsep{1pt}
\centering

\begin{tabular}{c c c c c c}
\rotatebox[origin=c]{90}{GT} &
\raisebox{-0.5\height}{\includegraphics[width=0.185\textwidth]{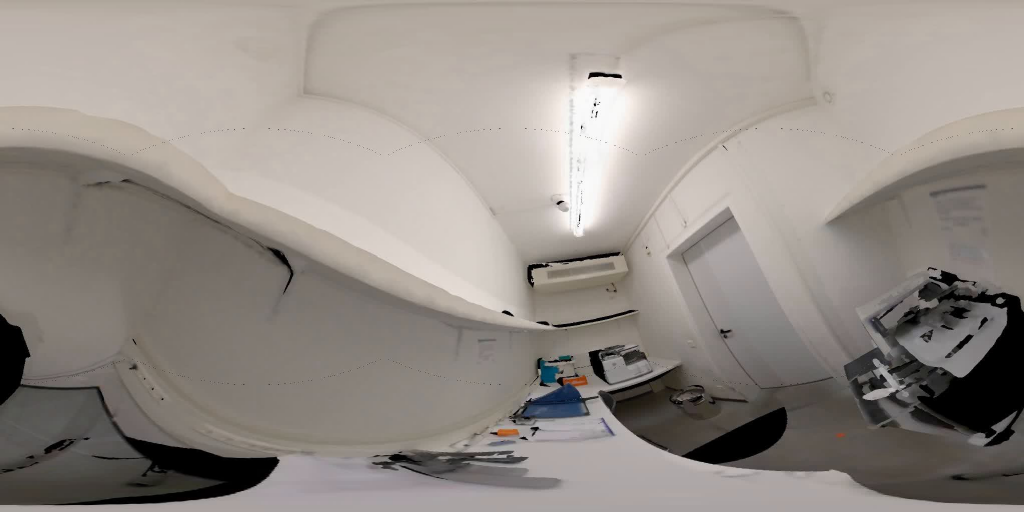}} &
\raisebox{-0.5\height}{\includegraphics[width=0.185\textwidth]{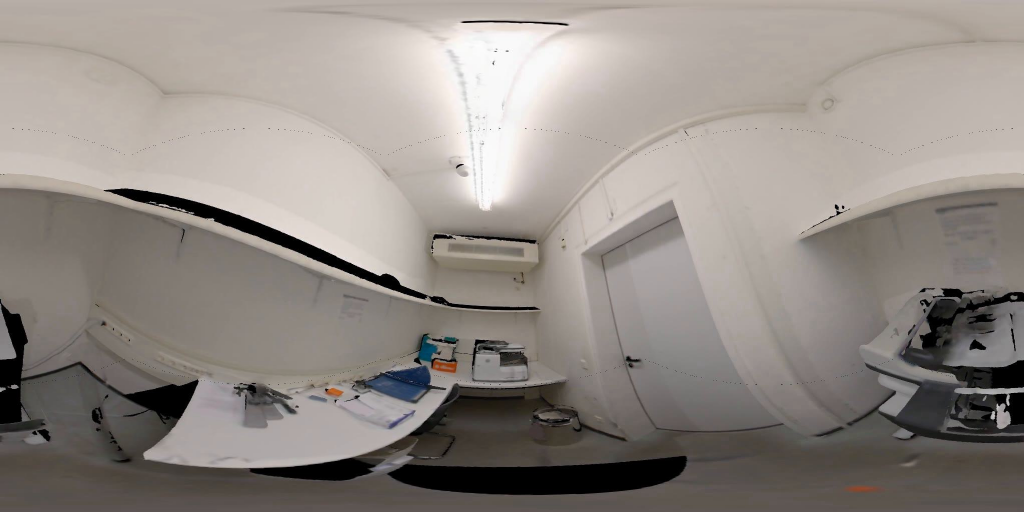}} &
\raisebox{-0.5\height}{\includegraphics[width=0.185\textwidth]{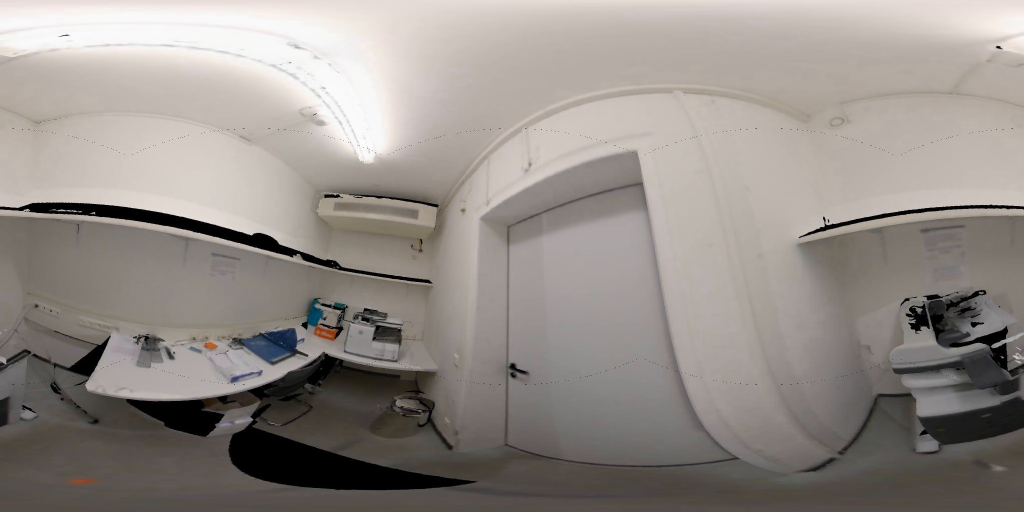}} &
\raisebox{-0.5\height}{\includegraphics[width=0.185\textwidth]{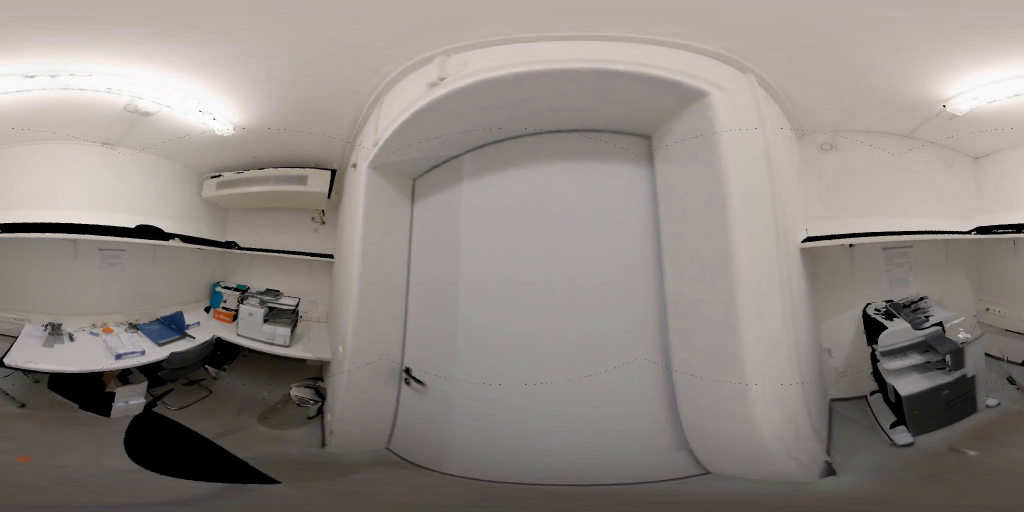}} &
\raisebox{-0.5\height}{\includegraphics[width=0.185\textwidth]{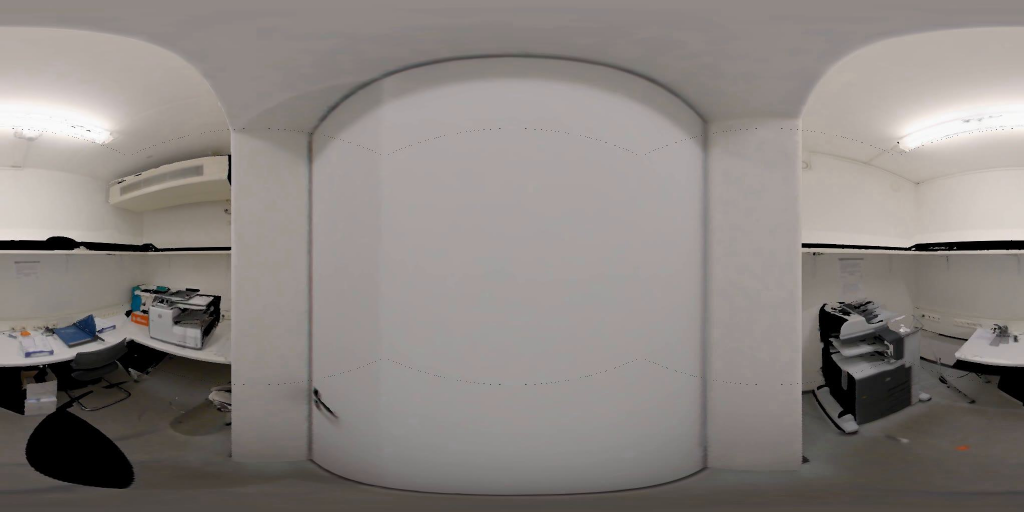}}\\
\rotatebox[origin=c]{90}{CogVideoX*} &
\raisebox{-0.5\height}{\includegraphics[width=0.185\textwidth]{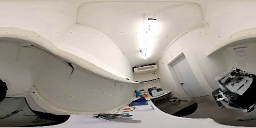}} &
\raisebox{-0.5\height}{\includegraphics[width=0.185\textwidth]{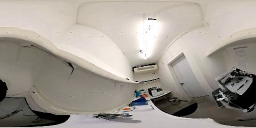}} &
\raisebox{-0.5\height}{\includegraphics[width=0.185\textwidth]{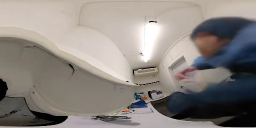}} &
\raisebox{-0.5\height}{\includegraphics[width=0.185\textwidth]{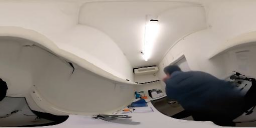}} &
\raisebox{-0.5\height}{\includegraphics[width=0.185\textwidth]{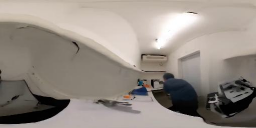}}\\
\rotatebox[origin=c]{90}{Wan2.2*} &
\raisebox{-0.5\height}{\includegraphics[width=0.185\textwidth]{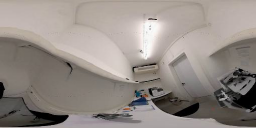}} &
\raisebox{-0.5\height}{\includegraphics[width=0.185\textwidth]{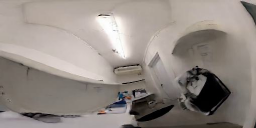}} &
\raisebox{-0.5\height}{\includegraphics[width=0.185\textwidth]{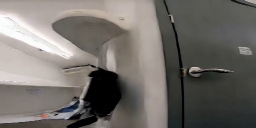}} &
\raisebox{-0.5\height}{\includegraphics[width=0.185\textwidth]{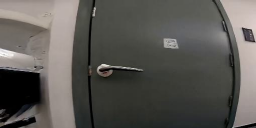}} &
\raisebox{-0.5\height}{\includegraphics[width=0.185\textwidth]{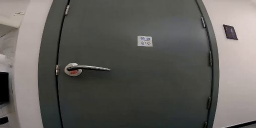}}\\
\rotatebox[origin=c]{90}{Hunyuan$^\dag$} &
\raisebox{-0.5\height}{\includegraphics[width=0.185\textwidth]{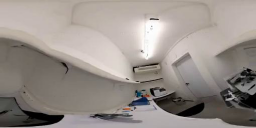}} &
\raisebox{-0.5\height}{\includegraphics[width=0.185\textwidth]{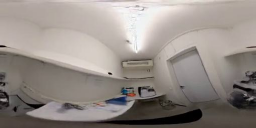}} &
\raisebox{-0.5\height}{\includegraphics[width=0.185\textwidth]{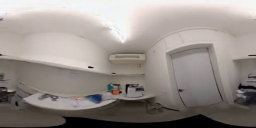}} &
\raisebox{-0.5\height}{\includegraphics[width=0.185\textwidth]{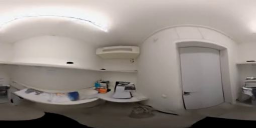}} &
\raisebox{-0.5\height}{\includegraphics[width=0.185\textwidth]{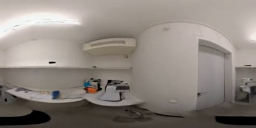}}\\
\rotatebox[origin=c]{90}{Wan2.1$^\dag$} &
\raisebox{-0.5\height}{\includegraphics[width=0.185\textwidth]{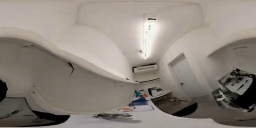}} &
\raisebox{-0.5\height}{\includegraphics[width=0.185\textwidth]{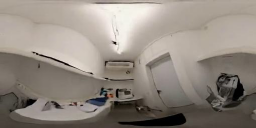}} &
\raisebox{-0.5\height}{\includegraphics[width=0.185\textwidth]{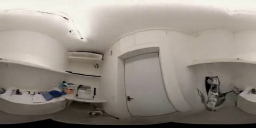}} &
\raisebox{-0.5\height}{\includegraphics[width=0.185\textwidth]{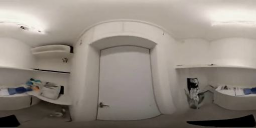}} &
\raisebox{-0.5\height}{\includegraphics[width=0.185\textwidth]{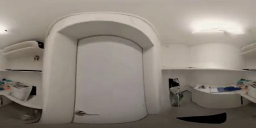}}\\
\rotatebox[origin=c]{90}{Wan2.2$^\dag$} &
\raisebox{-0.5\height}{\includegraphics[width=0.185\textwidth]{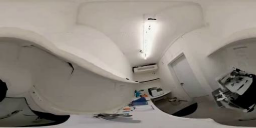}} &
\raisebox{-0.5\height}{\includegraphics[width=0.185\textwidth]{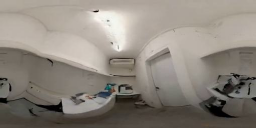}} &
\raisebox{-0.5\height}{\includegraphics[width=0.185\textwidth]{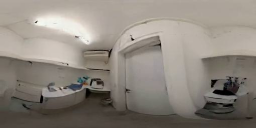}} &
\raisebox{-0.5\height}{\includegraphics[width=0.185\textwidth]{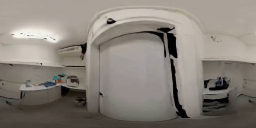}} &
\raisebox{-0.5\height}{\includegraphics[width=0.185\textwidth]{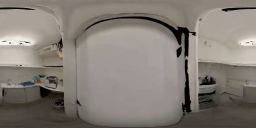}}\\
\rotatebox[origin=c]{90}{CogVideoX$^\dag$} &
\raisebox{-0.5\height}{\includegraphics[width=0.185\textwidth]{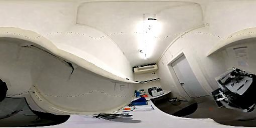}} &
\raisebox{-0.5\height}{\includegraphics[width=0.185\textwidth]{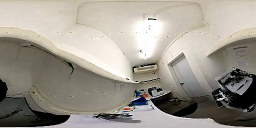}} &
\raisebox{-0.5\height}{\includegraphics[width=0.185\textwidth]{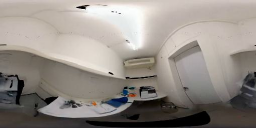}} &
\raisebox{-0.5\height}{\includegraphics[width=0.185\textwidth]{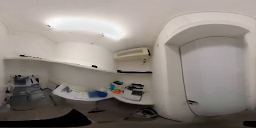}} &
\raisebox{-0.5\height}{\includegraphics[width=0.185\textwidth]{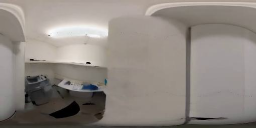}}\\
\end{tabular}
\caption{Qualitative results in ScanNet++ for the situation \textit{``If I want to stand by the gray door one meter ahead, with the wall-mounted air-conditioner one meter to the left.''}. * denotes prompt extension, while $\dag$ denotes fine-tuning.}
\label{fig:qualitative_results_scannetpp}
\end{figure*}

\begin{table}[]
    \centering
    \caption{Hyperparameters for fine-tuning the video generation models.}
    \small
    \resizebox{\textwidth}{!}{
    \begin{tabular}{lccc}
    \toprule
    Hyperparameter
    & Wan~\cite{wan2025wan} (LoRA)
    & HunyuanVideo~\cite{kong2024hunyuanvideo} (LoRA)
    & CogVideoX-1.5~\cite{yang2024cogvideox}(SFT) \\
    \midrule
    Optimizer              & AdamW             & AdamW             & AdamW              \\
    Weight decay           & $0$               & $0$               & $10^{-4}$               \\
    Betas                  & $[0.9,\,0.999]$   & $[0.9,\,0.999]$   & $[0.9,\,0.999]$    \\
    Learning rate          & $1\times10^{-4}$  & $1\times10^{-4}$  & $10^{-4}$              \\
    Warmup steps           & $0$               & $0$               & $500$              \\
    Number of workers      & $4$               & $4$               & $8$                \\
    Parallel strategy      & Accelerate DDP    & Accelerate DDP    & DeepSpeed ZeRO-2   \\
    Type of GPUs           & A100              & A100              & A100               \\
    Memory                 & 40\,GB            & 40\,GB            & 40\,GB             \\
    Number of GPUs         & $4$               & $4$               & $4$                \\
    Batch size per GPU     & $1$               & $1$               & $1$                \\
    Training precision     & bfloat16          & bfloat16          & bfloat16           \\
    Epochs                 & $8$               & $8$               & $10$               \\
    LoRA rank              & $32$              & $64$              & --                 \\
    \bottomrule
    \end{tabular}
    }
    \label{tab:training_config}
\end{table}

\begin{table}[!b]
    \centering
    \scriptsize 
    \caption{Per-class scores with Ground-Truth (GT) frames.}
    \resizebox{\textwidth}{!}{%
    \begin{tabular}{c|c|c|c|c|c|c|c|c|c|c|c|c|c|c}
    \toprule
         \multirow{3}{*}{Input}& \multicolumn{7}{|c}{HM3D}& \multicolumn{7}{|c}{ScanNet++}\\
         \cmidrule(l){2-15}
         &\multicolumn{4}{|c|}{Path $s_o \rightarrow s_T$} & \multicolumn{3}{|c}{End $s_T$} &\multicolumn{4}{|c|}{Path $s_o \rightarrow s_T$} & \multicolumn{3}{|c}{End $s_T$} \\
         \cmidrule(l){2-15}
         &LS&SE&OR&RP&OP&Aff.&ESR&LS&SE&OR&DC&OP&Aff.&ESR\\
    \midrule
    $s_o$&76.0&53.5&\textbf{50.8}&32.5&54.8&44.5&26.0&\textbf{23.7}&53.0&\textbf{30.1}&38.7&60.1&39.1&37.4\\
    $s_{\Delta5}$ (CogVideoX1.5)&\textbf{76.4}&57.8&41.6&34.1&57.6&45.4&29.1&23.5&\textbf{56.0}&29.6&\textbf{41.4}&60.5&42.6&39.5\\
    \rowcolor{gray!15}
    $s_{\Delta5}$ (GT)&73.2&\textbf{58.1}&40.4&\textbf{36.6}&\textbf{62.4}&\textbf{50.6}&\textbf{38.7}&22.2&55.4&28.8&40.8&\textbf{60.8}&\textbf{43.3}&\textbf{44.4}\\
    \bottomrule
    \end{tabular}
    }%

    \label{tab:gt}
\end{table}
\begin{table}[b]
    \centering
    \vskip-3.5ex
    \caption{Spearman correlation with human judgments across different LLM judges and prompt templates on 800 samples. And the agreement of different settings with our method.}
    \begin{tabular}{c|cc|ccc|c}
    \toprule
         \multirow{2}{*}{}& \multicolumn{2}{c|}{LLM judge} & \multicolumn{3}{c|}{Prompt Template}&\multirow{2}{*}{Ours}\\
         &Gemini~\cite{team2023gemini}& Claude~\cite{anthropic2024claude}& OpenEQA~\cite{OpenEQA2023} & MSQA~\cite{linghu2024multimodal_situated_3d}&No Verbosity Bias&\\
    \midrule     
    Human&0.806&0.865&0.957&0.918&0.914&\textbf{0.972}\\
    Ours&0.822&0.881&0.985&0.939&0.935&1\\
    \bottomrule
    \end{tabular}

    \label{tab:llmjudge_ablation}
\end{table}
\end{document}